\documentclass[10pt,twocolumn,letterpaper]{article}

\usepackage[pagenumbers]{cvpr} 

\usepackage[final]{graphicx}   
\usepackage{amsmath}
\usepackage{amssymb}
\usepackage{booktabs}
\usepackage{times}
\usepackage{epsfig}
\usepackage{arydshln}
\usepackage[dvipsnames]{xcolor}
\usepackage[toc,page]{appendix}

\usepackage[pagebackref,breaklinks,colorlinks]{hyperref}

\usepackage[capitalize]{cleveref}
\crefname{section}{Sec.}{Secs.}
\Crefname{section}{Section}{Sections}
\Crefname{table}{Table}{Tables}
\crefname{table}{Tab.}{Tabs.}

\def\cvprPaperID{1614} 
\def\confName{CVPR}
\def\confYear{2022}

\definecolor{ab}{rgb}{0.0, 0.58431372549019607843137254901961, 0.65098039215686274509803921568627}

\definecolor{note}{rgb}{0.81568627451,0.21568627451,0.21568627451}

\definecolor{newtext}{rgb}{0.0457464363464789,0.464567523657459760,0.1567890967856745}

\definecolor{yt}{rgb}{0.0457464363464789,0.464567523657459760,0.1567890967856745}

\definecolor{km}{rgb}{0.03921568627,0.43137254902,0.82352941176}

\begin{document}

\title{ScaleNet: A Shallow Architecture for Scale Estimation}

\author{
Axel Barroso-Laguna\hspace{50pt}
Yurun Tian\hspace{50pt}
Krystian Mikolajczyk\\
Imperial College London\\
{\tt\small \{axel.barroso17, yurun.tian, k.mikolajczyk\}@imperial.ac.uk}
}

\maketitle

\begin{abstract}

In this paper, we address the problem of estimating scale factors between images.  We formulate 
 the scale estimation problem as a prediction of a probability distribution over scale factors. 
 We design a new architecture, ScaleNet, that exploits dilated convolutions as well as self- and cross-correlation layers to predict the scale 
 between images. 
We demonstrate that  rectifying images with estimated scales leads to significant performance improvements for various tasks and methods. Specifically, we show how ScaleNet can be combined with sparse local features and dense correspondence networks to improve camera pose estimation, 3D reconstruction, or dense geometric matching in different benchmarks and datasets.
We provide an extensive evaluation on several tasks, and analyze the computational overhead of ScaleNet.
The code, evaluation protocols, and trained models are publicly available at \url{https://github.com/axelBarroso/ScaleNet}.

\end{abstract}

\section{Introduction}
Establishing correspondences is the very first step in many different 3D pipelines. Advancing on this task will have a direct impact on the performance of downstream applications such as camera pose estimation~\cite{sarlin2019coarse}, autonomous driving\cite{burki2019vizard}, or 3D reconstructions\cite{schonberger2016structure}. However, methods that search for correspondences between images face significant challenges, and although some solutions are more mature than others, the task still is far from being solved. \newline
\indent
As the field advances, even though the intermediate tasks in the correspondence search remain the same, their methods are being revisited and redesigned, \textit{e.g.}, keypoint detectors/descriptors~\cite{detone2018superpoint,dusmanu2019d2,revaud2019r2d2}, dense geometric matchers~\cite{Melekhov2018,GLUNet_Truong_2020},  or geometric verification techniques~\cite{goodcorr_kwang, acne_kwang, superglue2019}. These new approaches have shown that the downstream tasks can be pushed to new performance levels through robust correspondences. The key objective of these new methods is to handle more and more extreme cases where previous pipelines failed, and although some methods are arguably application-specific~\cite{zhou2017progressive}, their robustness to extreme conditions is the main reason for success.\newline
\indent
\begin{figure}[t]
  \hspace{-0.2cm}
 \centering
   \includegraphics[scale=0.4]{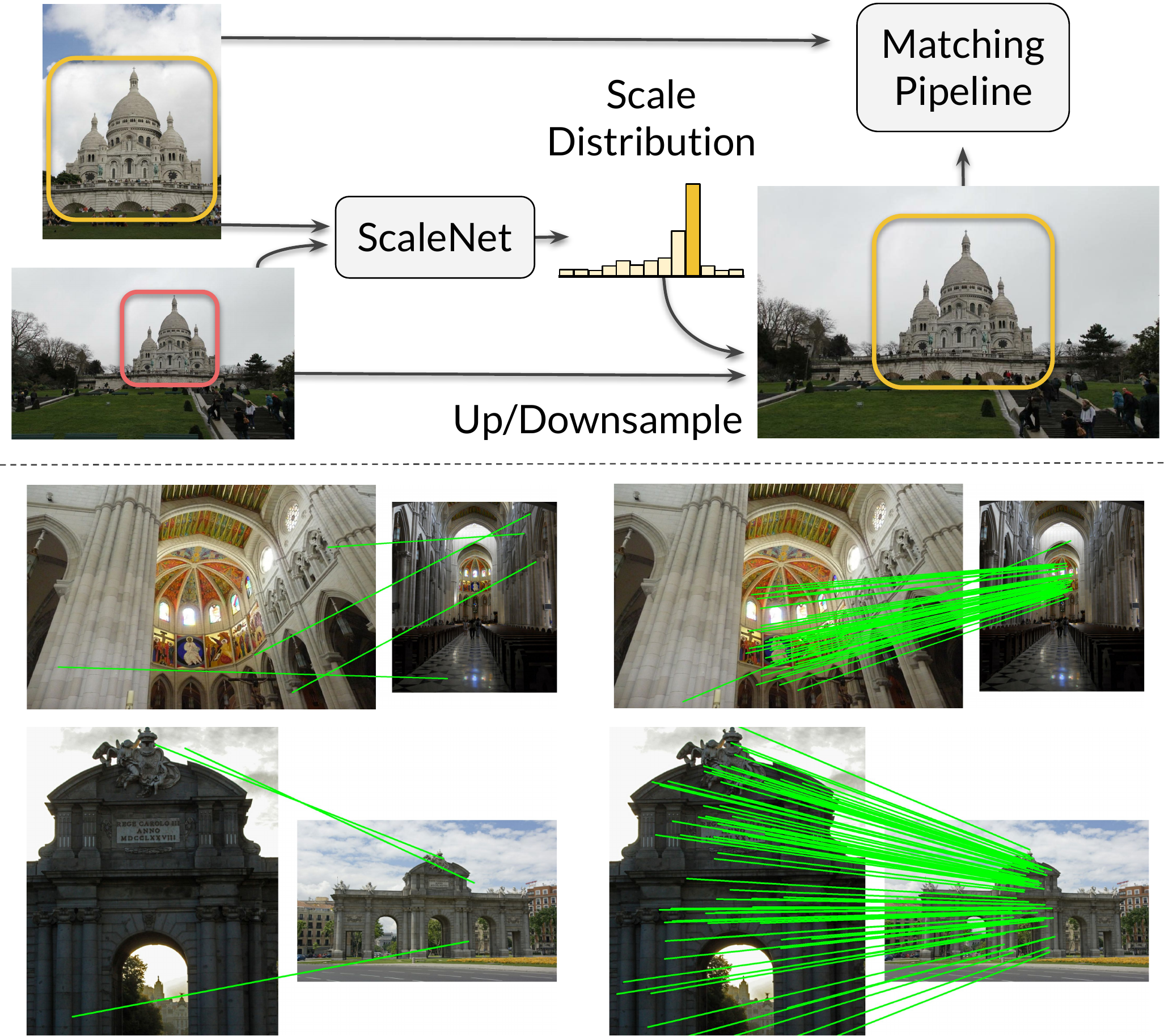}
  \vspace{-0.15cm}
    \caption{
    We propose a scale-aware system that helps establish correspondences under strong scale changes, allowing to correctly match pairs where previous pipelines were not successful. We introduce ScaleNet, a network that estimates the scale distribution between two images so that regions of interest can be corrected and exhibit the same scale factor (top). We display (bottom) an indoor and outdoor example where matches from R2D2 with multi-scale pyramid (left) increase if using ScaleNet rectification (right).}
    \label{fig:scalenet_start}
    \vspace{-0.25cm}
\end{figure}
Inspired by the previous methods targeting visual robustness \cite{yu2011asift, mishkin2015mods, mishkin2018repeatability}, we address  the problem of handling the scale change between images, which is a \mbox{long-standing} challenge in computer vision \cite{mikolajczykIJCV04,DoG}. Scale robustness and estimation have been the focus of much research in the area of handcrafted feature extraction \cite{TuytelaarsMikolajczyk2007,DoG,zhou2017progressive} as a reliable solution that can significantly boost the performance of existing methods. Moreover, the scale change is arguably the most challenging, and the most important parameter to estimate compared to rotation, translation, or even local affine deformations \cite{TuytelaarsMikolajczyk2007}. There are several strategies to deal with scale changes, with the multi-scale pyramid being one of the most popular solutions \cite{DoG, TuytelaarsMikolajczyk2007, dusmanu2019d2, revaud2019r2d2, hdd2020, laguna2019key, truong2021learning}. Although the multi-scale pyramid mitigates the problem of different scales, it increases complexity and ambiguity as the matcher needs to establish correspondences among multiple scale levels. 
Figure~\ref{fig:scalenet_start} (bottom) shows an example of extreme pairs for which R2D2\cite{revaud2019r2d2} with multi-scale pyramid can only get a high number of correct matches once we rectify the scales.
Besides the added complexity, a multi-scale pyramid is not always a straightforward solution to incorporate in some methods, such as dense correspondence networks \cite{GLUNet_Truong_2020, jiang2021cotr, sun2021loftr}.
In contrast to multi-scale pyramids, some works aim at being invariant to different scales through their learning process \cite{Pautrat_2020_ECCV, detone2018superpoint}, however, as a side effect, they become progressively less discriminative \cite{wiles2020d2d}. Another possible direction, and popular strategy, is to estimate the local or global transformations and rectify the images prior to establishing the correspondences\cite{yu2011asift, mishkin2018repeatability, toft2020singleimage,rau2020predicting}.

The scale factor characterizes the relationship between pairs of images and, in general, an accurate estimate can only be achieved when considering both images at the same time. Using pairs of images as input may increase the complexity beyond acceptable in some applications such as \mbox{large-scale} retrieval and localization unless used in their final verification stage.
Nonetheless, solving the scale before the main analysis improves the discriminative power and allows less robust but more efficient methods to be used in challenging scenarios \cite{Pautrat_2020_ECCV}. 
Hence, we propose a new approach that estimates and corrects the scale factor between a given pair of images before the correspondence search, which is illustrated at the top of figure \ref{fig:scalenet_start}. Our scale predictor network, termed ScaleNet, is inspired by dense geometric methods \cite{Melekhov2018, GLUNet_Truong_2020,truong2021gocor} and conditioned local features extractors \cite{wiles2020d2d, germain2020s2dnet, Pautrat_2020_ECCV}. ScaleNet extracts features from two low-resolution images and exploits CNN correlation layers to predict the scale factor. Due to the non-linear nature of scale changes, we formulate the scale regression problem as the estimation of a probability distribution in logarithmic space.
We show how ScaleNet can be combined with different methods and demonstrate the improvements on different tasks and datasets. 

Our contributions include: 1) a scale-aware matching system based on a novel scale predictor architecture, 2) a strategy to measure and label the scale factor between two images, and 3) a learning scheme that tackles the non-linear nature of scale changes.

\section{Related work}

Recent works have allowed significant progress in establishing good correspondences between images. Although many works have focused on solving entire tasks in an end-to-end manner \cite{sarlin2019coarse, brachmann2019neural}, there are lots of efforts focused on identifying limitations and improving the robustness of individual steps in modern pipelines \cite{mishchuk2017working, laguna2019key, superglue2019}.\\


\noindent
\textbf{Image rectification} consists of predicting or applying a set of transformations to the images so that the search for correspondences is done in an optimum setting. Pioneering work on this area is ASIFT \cite{yu2011asift}, which applies multiple affine transformations to find less challenging image pairs for matching with SIFT \cite{DoG}. 
MODS \cite{mishkin2015mods} investigated this line of work by introducing an iterative scheme to generate intermediate synthetic views between images. MODS also proposed an adaptive system to avoid applying synthetic transformations to easy-to-match pairs, being faster and more versatile than previous ASIFT. 
Closer to our work is \cite{zhou2017progressive}, which computes the scale factor between a pair of images by detecting and matching exhaustively SIFT features on multiple scale levels. Although it shows that they can deal with strong scale changes, their method is tied to the need of visiting all possible scaled images before the actual local feature matching stage.
One negative aspect of these methods is that they still require a blind exploration of synthetic views to find the optimal image transformations. A few works have tried to overcome the previous limitation and proposed to learn such parameterizations directly from the images. 
One of the first attempts is \cite{assignOrient2015}, where authors introduced a neural network to assign a canonical orientation to every input image patch before the descriptor architecture.
In AffNet \cite{mishkin2018repeatability}, authors follow this trend and learn a full affine shape estimator to geometrically align input patches before the descriptor. 
Moreover, \cite{rau2020predicting} proposed a scene-specific overlapping estimator and showed that scale rectification based on image overlapping can improve feature matching.
Unlike previously learned methods, our labeling strategy is based on keypoint distance ratios, resulting in a scene-agnostic scale predictor that uses pairs of images as input.
Even though our ScaleNet tackles only the scale factor out of several possible transformation parameters, we show in section \ref{Sec:Experiments} that the scale factor is crucial for boosting the performance of current methods.\\

\begin{figure*}[t]
 \centering
   \includegraphics[scale=0.72]{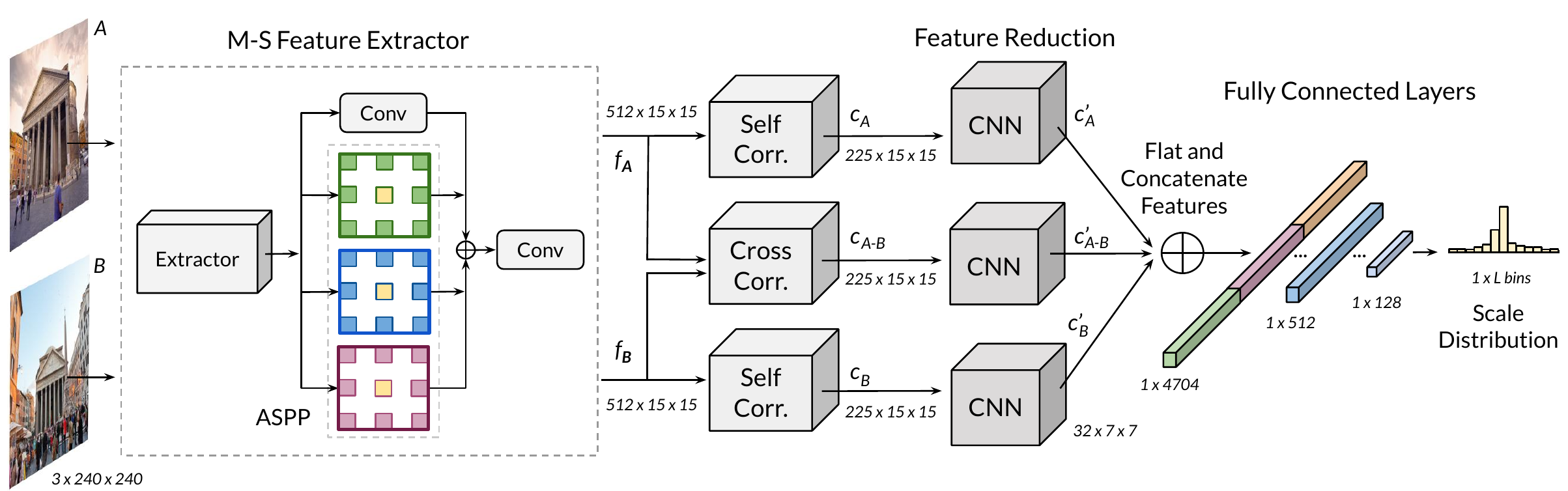}
    \caption{ScaleNet uses a multi-scale (M-S) feature extractor and a combination of self- and cross-correlation layers to calculate the relationship within each image and between them. The M-S extractor is composed of a common CNN and an ASPP block \cite{chen2017deeplab, chen2017rethinking}. 
    The ASPP module consists of three 3x3 convolutional layers with different dilation rates. 
    \textit{Conv} blocks perform 1x1 convolutions to learn the cross-channel interactions that go into the self- and cross-correlation layers. Correlation volumes' dimensionality is reduced through a CNN, and its results are concatenated into a flat feature vector. Finally, consecutive fully connected layers calculate the scale distribution.}
    \label{fig:scalenet_architecture}
\end{figure*}
\noindent 
\textbf{Visual robustness}  has been the focus of numerous works in the field of correspondence search \cite{LuoASLFeat2020, ebel2019beyond, hdd2020}. The rotation has been addressed by correcting input patches \cite{DoG, ebel2019beyond, mishkin2018repeatability} before extracting local descriptors \cite{mishchuk2017working,tian2019sosnet,tian2020hynet}, or by designing robust architectures \cite{LuoASLFeat2020, hdd2020, liu2019gift}. However, in the context of scale changes, the standard strategy is the well-known multi-scale (M-S) pyramid approach, which applies methods at different re-scaled versions of the image \cite{revaud2019r2d2, dusmanu2019d2,d2d2020}. Even though M-S pyramids offer a versatile solution for many applications, it does not provide a suitable approach for extreme scale changes \mbox{(cf. section \ref{Sec:Experiments})}, and thus, both, single and multi-scale feature extractors, benefit from correcting the scale factor before extracting features.
Moreover, recent works show that there is growing interest in pair-wise methods, \textit{i.e.}, that use two input images at the same time to establish the local or dense correspondences \cite{sun2021loftr, jiang2021cotr, germain2020s2dnet, Melekhov2018, GLUNet_Truong_2020, wiles2020d2d, pautrat2020online}, but, in that scenario, there is no clear or effective strategy for dealing with scale changes. Hence, given that such methods already take two images as inputs, ScaleNet rectification offers a more natural and intuitive process towards visual robustness than M-S pyramids.

\section{Method}
\label{sec:method}
ScaleNet embraces several key concepts to deliver good performance in practical settings. The first key aspect is that it is effective for low-resolution images, which makes it more efficient. Another important idea is the formulation of the scale estimation as a distribution prediction in logarithmic space rather than a regression problem \cite{mishkin2018repeatability, yi2016lift, jaderberg2015spatial}. This allows using a simple and shallow, yet effective architecture.  Figure \ref{fig:scalenet_architecture} presents our ScaleNet architecture, and the following sections detail each of the aspects of ScaleNet and its learning scheme.

\subsection{ScaleNet architecture}
Consider $A$ and $B$ as input images to ScaleNet. 
Low-resolution images $A$ and $B$  are processed by a multi-scale feature extractor block, which is composed of a generic network, \textit{e.g.}, VGG, or ResNet, followed by the atrous spatial pyramid pooling (ASPP)  \cite{chen2017deeplab, chen2017rethinking}. 
The ASPP block achieves multi-scale robustness by applying to the feature map $3\times3$ dilated convolutions, each with a different dilation rate, \textit{e.g.}, 2, 3, and 4. Thus, the ASPP block allows the network to compute and fuse features from different receptive fields. M-S features are concatenated and fed into a final $1 \times 1$ convolution, which combines the features from local and global areas at a minimum cost.
We then apply self- and cross-correlation layers to multi-scale features $f_A$ and $f_B$, and obtain the correlation maps, $c_A$, $c_B$, and $c_{A-B}$, which contain the self- and cross-pairwise similarities. As in \cite{Rocco17}, ReLU and L2-normalization are applied to the correlation maps before the feature reduction blocks, which are composed of four Conv-Batch-ReLU layers each. Finally, $c'_A$, $c'_B$, and $c'_{A-B}$ maps are flattened, concatenated, and fed into a set of fully connected layers to predict the scale \mbox{distribution $P$}, with a final softmax activation layer. 

\subsection{Predicting scale distributions}
\label{Predicting_Scale}
ScaleNet outputs a scale distribution rather than a regressed single scale factor, which helps the network to converge to a reliable model. 
In contrast, when tackling the problem as a regression task, the same network cannot predict an accurate scale (cf. appendix \ref{appendix_sec:experiments_ablation}). 
We attribute this to the fact that the network can learn and interpret the relationships between the quantized scale ranges and solve an easier classification task, which requires it to assign the weights to the predefined scale factors instead of predicting its actual value.\\
\indent
We formulate the scale estimation as a problem of predicting the probability distribution in a scale-space quantized into $L$ bins. Given images $A$ and $B$, our objective function measures the distance between our computed scale distribution, $P_{A\rightarrow B}$, and the ground-truth distribution $P_{A\rightarrow B}^{gt}$:
\begin{equation}
\begin{gathered}
    Loss(A, B) = KL(P_{A\rightarrow B}, P_{A\rightarrow B}^{gt}),
\label{eq:loss}
\end{gathered}
\end{equation}
where 
$KL(\cdot, \cdot)$ 
is the Kullback-Leibler divergence loss. To obtain the scale factor from the probability distribution $P_{A\rightarrow B}$, we combine all scale levels using a soft-scale computation. It enables the network to output scale factors that interpolate  between the quantized scale values, thus covering all possible scales between images $A$ and $B$. \mbox{Soft-assignment} gives the architecture further flexibility and robustness when inferring scales as shown in section \ref{sec:3d_results}. \\
\indent
The scale factor is a relative ratio operator, hence, it is \mbox{non-linear}. To avoid a bias by high scale values when computing the soft-scale, we transform the quantized scale classes, $s_i$, to logarithmic space. The logarithmic transformation allows to calculate the soft-scale, $\bar{S}_{A\rightarrow B}$, as a linear combination of the logarithmic scales, $\bar{s}_i$, weighted by the predicted scale probabilities from softmax output $p_i$. The global scale factor $\bar{S}_{A\rightarrow B}$ in log-scale is given as:
\begin{equation}
\begin{gathered}
    \bar{S}_{A\rightarrow B} = \sum_{i=0}^{L-1} p_i \cdot ln(s_i),
\label{eq:scale}
\end{gathered}
\end{equation}
 where $s_i=\sigma^t$ corresponds to the quantized scale for bin $i$,  
$\sigma$ is our predefined base scale factor, integer \mbox{$t\in[-(L-1)/2,\ldots,0,\ldots,(L-1)/2]$}, and  $L$ is the total number of scale bins. 
Moreover, we improve the robustness of our scale estimator by a simple yet effective consistency check trick, where we compute the scale factor, $\bar{S}_{A\rightarrow B}$, and its inverse, $\bar{S}_{B\rightarrow A}$, and combine them as:
\begin{equation}
\begin{gathered}
  \hat{S}_{A \rightarrow B} = \frac{\bar{S}_{A \rightarrow B} - \bar{S}_{B \rightarrow A}}{2} \quad \textrm{and} \quad S_{A \rightarrow B} = e^{\hat{S}_{A \rightarrow B}},
\label{eq:cyclic}
\end{gathered}
\end{equation}
with $S_{A\rightarrow B}$ as the final scale factor between the images.

\subsection{Dataset generation}
\label{dataset_generation}
\begin{figure}[t]
 \centering
   \includegraphics[scale=0.4]{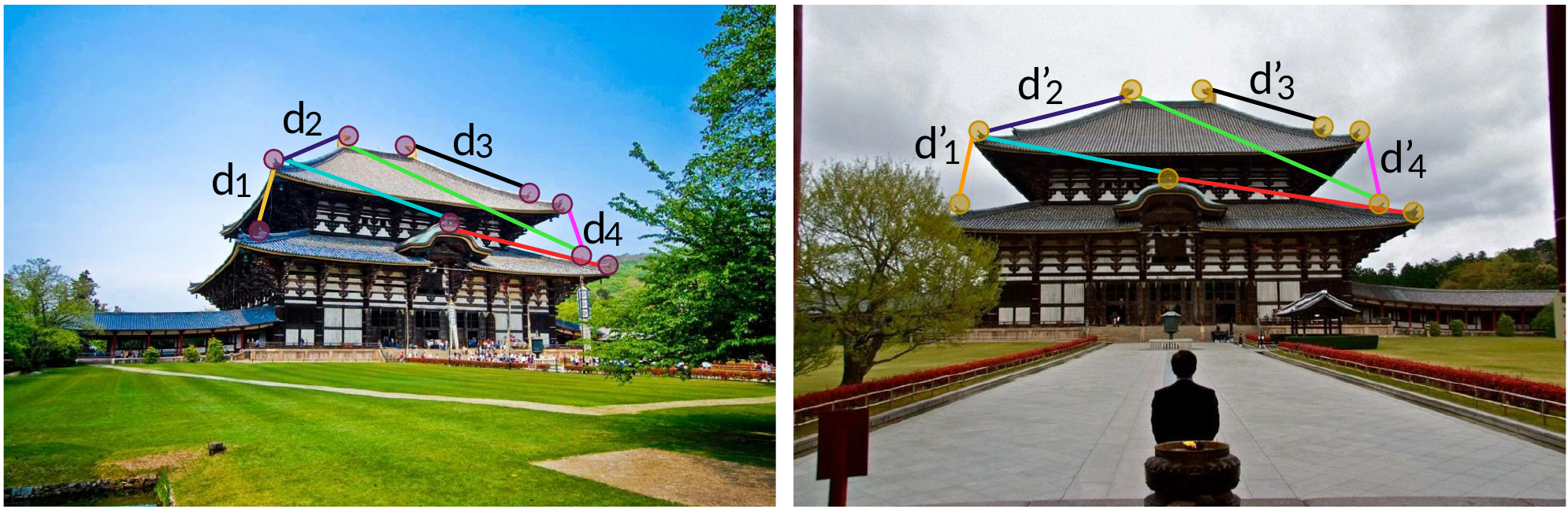}
    \caption{In the dataset generation, we randomly pick pairs of corresponding points in both views and compute their distances in the image planes. The final scale ground-truth is the average of the distance ratios of all picked keypoints.}
    \label{fig:dataset}
\end{figure}
ScaleNet is trained with synthetically generated image pairs as well as images from real scenes. \\

\noindent
\textbf{Synthetic pairs.} We define a set of planar affine transformations to map one image into another. Such image pairs are easy to generate on-demand for any ground-truth scale $S_{gt}$, however, they do not include the real noise from different imaging conditions. \\ 

\noindent
\textbf{Real pairs} present more challenging conditions than synthetic pairs, \textit{i.e.}, non-planar viewpoint changes, weather/illumination conditions, or occlusions, among others.
In addition, in contrast to the previous synthetic global transformations, the scale between two real images may spatially vary, and areas of different depth or strong perspective changes may include various scale factors. 
Thus, we introduce a new approach for obtaining training data by estimating the scale factors between real images. 
We use 3D reconstruction datasets, where 3D point clouds and their corresponding projected 2D positions on the images are available. 
First of all, given the 3D model, we find pairs of images with an overlap higher than 10\% computed as in \cite{dusmanu2019d2, ono2018lf}. 
For each pair of images $A$ and $B$, we identify the 3D points of the model that are visible in both images. Using the registered 3D points from the model as opposed to sampling random positions ensures that the regions used for computing the scale are discriminative. Thus, given the covisible 3D points, we query its projected 2D positions, $k_A$, and $k_B$, on image $A$ and $B$.
We randomly sample pairs of points, $k_A$ and $k_B$, and compute their distances as shown in figure \ref{fig:dataset}. 

The scale factor between two images with keypoints \mbox{$i$ and $j$} is calculated as the ratio of their distances in image $A$ and $B$:
\begin{equation}
\begin{gathered}
    S_{i-j} = \frac{\|k_{Bi}-k_{Bj}\|}{\|k_{Ai}-k_{Aj}\|} \quad \textrm{and} \quad i, j \in [1, ..., K], 
\label{eq:scale_ij}
\end{gathered}
\end{equation}
 with $K$ as the total number of covisible 3D points between images $A$ and $B$. As different regions may exhibit different scale factors,  we compute the global scale factor as the average of ratios in logarithmic scale after sampling $R$ different pairs of points:
\begin{equation}
\begin{gathered}
    S_{A\rightarrow B}^{gt} = e^{\bar{S}_{A\rightarrow B}} \quad \textrm{where} \quad \bar{S}_{A\rightarrow B} = \frac{1}{R}\sum_{i \neq j} ln(S_{i-j}).
\label{eq:scale_groundT}
\end{gathered}
\end{equation}
As detailed in section \ref{Predicting_Scale}, the ScaleNet learning scheme minimizes the K-L divergence between the predicted scales and ground-truth distributions. We, therefore, build the ground-truth scale distribution, $P_{A\rightarrow B}^{gt}$, such as it satisfies: 
\begin{equation}
\begin{gathered}
    ln(S_{A\rightarrow B}^{gt}) = \bar{S}_{A\rightarrow B}^{gt} = \sum_{i=0}^{L-1} p_i^{gt} * ln(s_i),  
\label{eq:scale_groundT2}
\end{gathered}
\end{equation}
where $s_i$ are the quantized scale factors, and $ P_{A\rightarrow B}^{gt} = [p_0^{gt}, ..., p_{(L-1)}^{gt}]$ is the ground-truth distribution of the scale estimated as a normalized histogram of point pairs.

\section{Implementation notes}

\noindent
\textbf{ScaleNet details. }
ScaleNet predicts a scale distribution with $L = 13$ bins and  
$\sigma = \sqrt{2}$, giving possible scale factors in range $S \in [0.16, ..., 6]$. ASPP module has 3 levels with dilation rates 2, 3, and 4. During training,
ScaleNet uses Adam Optimizer with a learning rate of $10^{-4}$ and a decay factor of 0.1 every ten epochs. The training takes on average 40 epochs, 20 hours on a machine with an i7-7700 CPU running at 3.60GHz, and an NVIDIA GeForce GTX \mbox{1080-Ti}. ScaleNet model and training scripts are implemented in \mbox{PyTorch \cite{paszke2017automatic}}.\\

\noindent
\textbf{Dataset details. }
We use Megadepth dataset \cite{li2018megadepth} for generating our custom training and testing dataset. We discard scenes that are in the PhotoTourism test from our training set as in \cite{superglue2019} to avoid overlap. We keep 10\% of the training scenes as our validation set.
During data generation, we sample $R=200$ random pairs of 3D points \mbox{(cf. equation \ref{eq:scale_groundT})} for a robust scale estimation between the two images.
We sample pairs of images with scale factors $S \in [0.16, ..., 6]$ and create a collection of 250,000 training and 25,000 validation pairs where all scale factors are well represented. Synthetic pairs are generated on the fly from the Megadepth training images during training. We include more details and examples of our training set in appendix \ref{appendix:implementation}.

\section{Experiments}
\label{Sec:Experiments}
This section presents results for ScaleNet integrated with state-of-the-art methods on several datasets and tasks. Refer to appendix for more experiments and qualitative examples.

\subsection{Preliminaries}
\label{Sec:Experiments_ablation}
\begin{figure}

\begin{subfigure}{.48\textwidth}
  \includegraphics[scale=0.36]{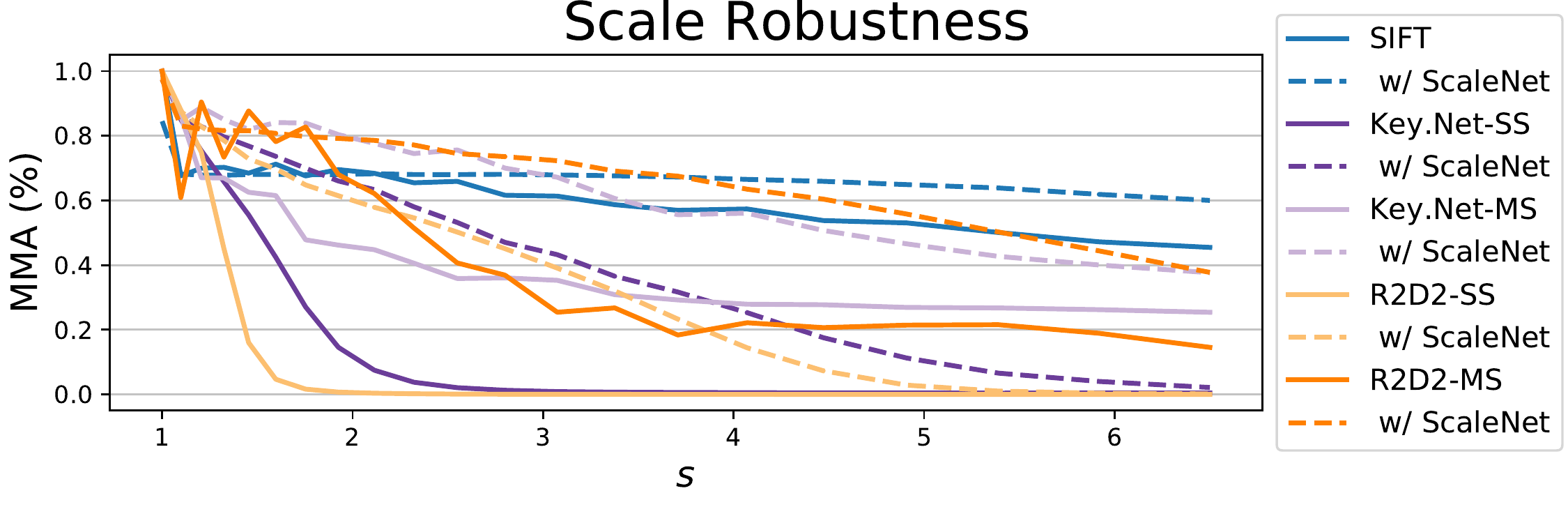}
  \caption{Mean matching accuracy under synthetic scale transformations, $s$.}
  \label{fig:mma_synthetic}
\end{subfigure}%

\vspace{0.05cm}
\begin{subfigure}{.47\textwidth}
  \hspace{-0.50cm}
  \includegraphics[scale=0.4]{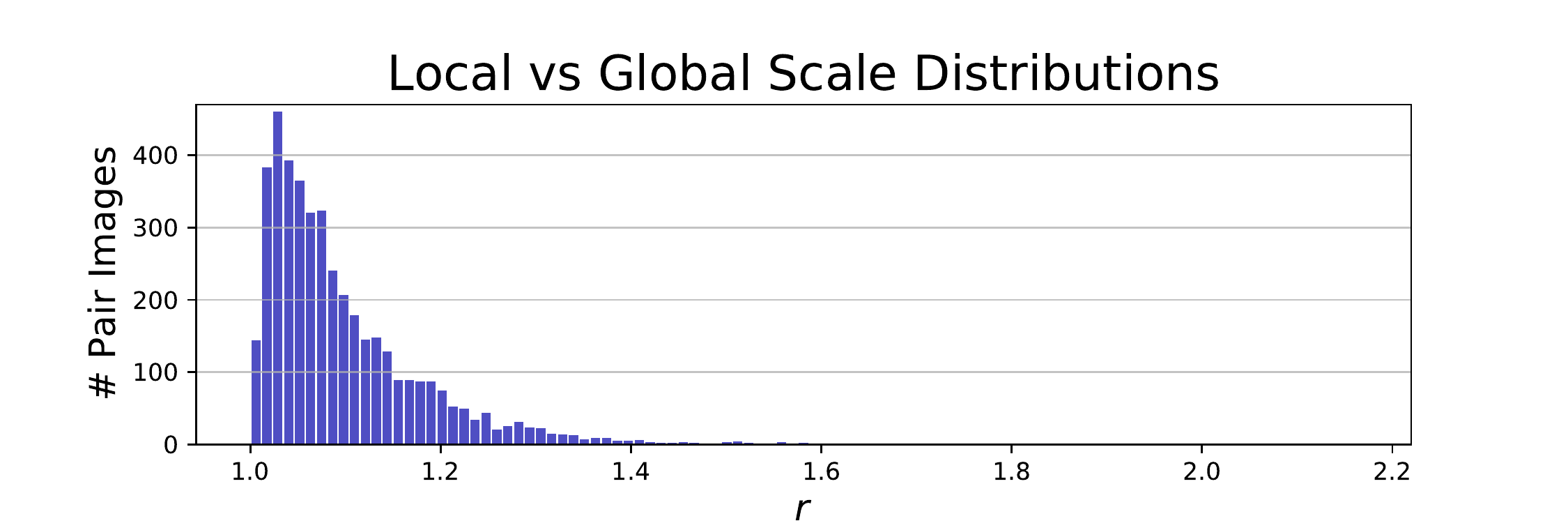}
  \caption{Histogram of the scale ratios, $r$, between the globally and locally estimated scales in image pairs used for training, with $r = s_a / s_b$, and where $s_a = max(s_{local}, s_{global})$, and $s_b = min(s_{local}, s_{global})$.}
  \label{fig:ablation_global}
\end{subfigure}
\vspace{0.10cm}
\caption{Analysis on the visual robustness against pure scale changes, and comparison of local vs global scales.}
\label{fig:ablation_fig}
\end{figure}

\begin{table}[]
\footnotesize
\begin{center}
\begin{tabular}{c c c c c}
\multicolumn{1}{c}{} & \multicolumn{3}{c}{\textbf{Pose estimation (AUC)}} & \multicolumn{1}{c}{\textbf{Time}}\\ 
\cline{2-4} 
\noalign{\smallskip}
 & at 5\textdegree & at 10\textdegree & at 20\textdegree & (ms)\\
\hline \noalign{\smallskip}
Baseline & 4.8 & 7.4 & 10.4 & - \\
\hdashline\noalign{\smallskip}
VGG-16 & 5.4 & 8.3 & 11.8 & 8.3 \\
ResNet-50 & 5.6 & 8.8 & 12.7 & 15.8  \\
\hdashline\noalign{\smallskip}
VGG w/ self-corr. & 6.2 & 9.3 & 12.5 & 11.3 \\
VGG w/ ASPP & 7.3 & 11.1 & 15.7 & 16.6 \\
\textbf{Ours} (VGG+self-corr+ASPP) & 8.4 & 12.3 & 17.6 & 19.4 \\
\textbf{Ours} + Consistency check & 8.7 & 13.4 & 19.5 & 19.8 \\
\end{tabular}
\end{center}
\normalsize
\vspace{-0.30cm}
\caption{Ablation study of the different ScaleNet's design choices. Baseline refers to SuperPoint \cite{detone2018superpoint} without scale correction.}
\label{tab:ablation_study}
\end{table}
\noindent
\textbf{Multi-scale pyramid \& ScaleNet.} 
Multi-scale (M-S) pyramids and ScaleNet aim at making methods more robust against arbitrary scale changes. Although both approaches tackle the same problem, each has its strengths, \textit{e.g.}, M-S pyramids can compute a higher number of features by visiting multiple resized images, and ScaleNet offers a more natural integration into tasks where two images are given as input \cite{ germain2020s2dnet, Melekhov2018, GLUNet_Truong_2020, wiles2020d2d}.
Besides, we claim that ScaleNet improves not only single-scale feature extractors but also multi-scale ones. 
We analyze in figure \ref{fig:mma_synthetic} the robustness of methods against synthetic scale transformations and show how the combination with ScaleNet benefits them. In this experiment, we use 2,000 random images from the Megadepth dataset and scale them to create the pairs. We measure the mean matching accuracy (MMA) computed as in \cite{revaud2019r2d2}. 
As expected, results indicate that the performance of single-scale methods, Key.Net\cite{laguna2019key} and R2D2\cite{revaud2019r2d2}, drop notably even when images present small scale perturbations ($s>1.5$). Meanwhile, M-S pyramid or ScaleNet strategies mitigate the effect and lead to a better approach. Moreover, we observe that the combination of M-S pyramids and ScaleNet achieves the top performance, and proves that both strategies contribute and work well together.\\


\noindent
\textbf{Local vs global scale estimation.} ScaleNet can predict a scale  for each point within the image, however,  it is not straightforward to correct the scale factor locally for networks that process the whole image, \textit{e.g.}, dense correspondences networks \cite{Melekhov2018,GLUNet_Truong_2020,truong2021gocor}, or dense local feature extractors \cite{detone2018superpoint,dusmanu2019d2,revaud2019r2d2}. Figure \ref{fig:ablation_global} shows the histogram of the average scale ratios between the globally and the locally estimated scales per image. To compute the local scale values, we restrict the random sampling to spatially neighboring keypoints in equation \ref{eq:scale_ij} rather than points sampled across the whole image. Figure \ref{fig:ablation_global} shows that in the majority of images the differences between global and local estimations are small and within 1.0 and 1.2. A ratio $r$ of 1.0 indicates that the local and the global scales between the two images are the same, and therefore, the global scale is valid across the whole image. Based on results in figure \ref{fig:ablation_global}, we argue that although local scales could bring an extra benefit in scenes with strong viewpoint or perspective changes, a single global scale will significantly contribute towards correcting images.\\

\begin{table}[]
\footnotesize
\begin{center}
\begin{tabular}{c c c c}
\multicolumn{1}{c}{} & \multicolumn{3}{c}{\textbf{Pose estimation (AUC)}}\\ 
\cline{2-4} 
\noalign{\smallskip}
 & at 5\textdegree & at 10\textdegree & at 20\textdegree \\
\hline \noalign{\smallskip}
SIFT & 9.8 & 15.4 & 22.0 \\
w/ ScaleNet & \textbf{12.9} (+32\%) & \textbf{19.5} (+27\%) & \textbf{27.4} (+25\%)\\
\hdashline\noalign{\smallskip}
R2D2-SS & 3.1 & 4.4 & 5.6 \\
w/ ScaleNet & \textbf{6.5} (+110\%) & \textbf{9.7} (+121\%) & \textbf{12.8} (+129\%)\\
\hdashline\noalign{\smallskip}
R2D2-MS & 10.5 & 13.1 & 18.9 \\
w/ ScaleNet & \textbf{13.2} (+26\%) & \textbf{17.5} (+34\%) & \textbf{21.0} (+11\%)\\
\hdashline\noalign{\smallskip}
Key.Net-SS & 4.7 & 6.9 & 9.3 \\
w/ ScaleNet & \textbf{8.9} (+89\%) & \textbf{14.0} (+103\%) & \textbf{19.5} (+110\%)\\
\hdashline\noalign{\smallskip}
Key.Net-MS & 14.0 & 22.1 & 31.6 \\
w/ ScaleNet & \textbf{17.2} (+23\%) & \textbf{26.4} (+20\%) & \textbf{38.4} (+22\%)\\
\hdashline\noalign{\smallskip}
SuperPoint & 5.4 & 8.2 & 11.0 \\
w/ ScaleNet & \textbf{8.1} (+50\%) & \textbf{12.4} (+51\%) & \textbf{17.3} (+57\%)  \\
\hdashline\noalign{\smallskip}
SP+SuperGlue & 16.5 & 25.3 & 35.0 \\
w/ ScaleNet & \textbf{22.5} (+36\%) & \textbf{33.9} (+34\%) & \textbf{45.5} (+30\%)\\
\end{tabular}
\end{center}
\normalsize
\vspace{-0.30cm}
\caption{Relative camera pose results on custom Megadepth split with and without ScaleNet correction.}
\label{tab:camera_pose_megadepth}
\end{table}

\noindent
\textbf{Ablation study} in table \ref{tab:ablation_study} displays the contributions of each ScaleNet's block towards robust scale estimation.
We combine different architecture's designs with SuperPoint\cite{detone2018superpoint} and test them in the task of relative camera pose estimation. We use Lowe’s ratio test\cite{DoG} and MAGSAC\cite{barath2019magsac} to compute camera poses, and, as in \cite{superglue2019, sun2021loftr}, report the AUC of the pose errors at $5^{\circ}$, $10^{\circ}$, and $20^{\circ}$, where the error is calculated as the maximum of the rotation and translation angular errors. We sample 2,000 image pairs with scale factors between 0.16 and 6 from an independent validation set. Besides the AUC, we report the overhead inference time of each design.\\
\indent
We first compare the effect of different pre-trained feature extractors on ImageNet \cite{imagenet_cvpr09} (cf. figure~\ref{fig:scalenet_architecture}), VGG-16 \cite{Simonyan15}, and ResNet-50 \cite{he2016deep}, and see that the more complex ResNet representation contributes towards better poses with the downside of a higher computational cost. 
Thus, to keep our method light and fast, and given the similarity of the AUC scores, we use the VGG feature extractor for following experiments.
In addition to the extractors, we analyze the effect of the self-correlation layers and the multi-scale ASPP component and observe that both boost the performance. Self-correlations capture the intra-image relationships and, therefore, give a better awareness of the global content. Additionally, ASPP offers a mechanism to extract more global features, thus, address larger scale changes. 
Furthermore, we show that the consistency check offers a higher AUC at a low computation cost. Note that M-S features and correlation layers only need to be computed once in inference and, hence, the extra cost for consistency check is small and proportional to running the dense layers twice.

\subsection{Relative camera pose}
\label{sec:camera_pose_megadepth}
\noindent
\textbf{Protocol.} We first evaluate ScaleNet on the camera pose estimation task due to its natural integration into the existing pipelines. Similar to the previous ablation experiment, given a collection of image pairs, we calculate the AUC of the camera pose error at $5^{\circ}$, $10^{\circ}$, and $20^{\circ}$ as in \cite{superglue2019, sun2021loftr}. We use the test scenes from Megadepth and mine 4,000 images pairs with small and strong scale changes such as $s \in [0.16, 6]$. Appendix contains more details and examples of our dataset. We study the effect of ScaleNet on popular and publicly available local feature methods \cite{detone2018superpoint, laguna2019key, mishchuk2017working, revaud2019r2d2, DoG}, and refer to \mbox{Key.Net/HardNet} as \mbox{Key.Net} in the following tables and figures.\\

\noindent
\textbf{Results} in table \ref{tab:camera_pose_megadepth} show that ScaleNet corrections boost the performance of all methods. As discussed previously, even though ScaleNet excels when combined with a single-scale method, \textit{e.g.}, SuperPoint, ScaleNet is also able to improve the pose estimation of multi-scale extractors. We observe that the average improvements are 77\% and 24\% for single and multi-scale methods, respectively. In addition, we report results with SuperPoint and SuperGlue \cite{superglue2019} and see that even this state-of-the-art matcher benefits (+33\% on average) from a scale correction prior to the feature extraction. 

\subsection{Geometric matching}
\label{sec:geometric_matching}

\begin{table}[]
\footnotesize
\begin{center}
\begin{tabular}{c c c c}
\multicolumn{1}{c}{} & \multicolumn{3}{c}{\textbf{Geometric matching (PCK-5 \%)}}\\ 
\cline{2-4} 
\noalign{\smallskip}
& All & Easy & Hard \\
\hline \noalign{\smallskip}
DGC-Net & 40.2 & 34.4 & 4.5 \\
w/ ScaleNet & \textbf{41.4} (+3\%) & \textbf{36.8} (+7\%) & \textbf{20.1} (+347\%) \\
\hdashline\noalign{\smallskip}
GLU-Net & 55.5 & 55.4 & 10.8 \\
w/ ScaleNet & \textbf{57.8} (+4\%) & \textbf{56.3} (+2\%) & \textbf{26.8} (+148\%)\\
\end{tabular}
\end{center}
\normalsize
\vspace{-0.30cm}
\caption{Results on sparse correspondences for full, easy, and hard splits in Megadepth, consisting of 1,600, 627, and 440 pairs, respectively. Improvements from ScaleNet are across all splits, in particular, geometric matchers benefit largely from scale correction when there are large scale changes ($s>1.8$).}
\label{tab:dense_matching_megadepth_v2}
\end{table}

\begin{table*}[!t]
\footnotesize
\begin{center}
\begin{tabular}{c c c c c c c c c c c c }

&  & \bf{Reg.} & \bf{Num.} & \bf{Track} & \bf{Rep.} & & \bf{Reg.} & \bf{Num.} & \bf{Track} & \bf{Rep.} \\
&  & \bf{Images} & \bf{Obs.} & \bf{Length} & \bf{Error} & & \bf{Images} & \bf{Obs.} & \bf{Length} & \bf{Error}  \\
\cline{2-11}  \noalign{\smallskip}

& SIFT & 98.5 & 93k & 7.26 & \textbf{0.80} & SuperPoint & 99.0 & 78k & 10.47 & \textbf{1.17} \\
& w/ ScaleNet& \textbf{98.6} & \textbf{126k} & \textbf{10.20} & 0.87 & w/ ScaleNet & \textbf{99.5} & \textbf{135k} & \textbf{12.71} & 1.26 \\
& w/ D-ScaleNet & 98.5 & 118k & 9.55 & 0.85 & w/ D-ScaleNet & 99.3 & 98k & 11.96 & 1.22 \\
\cdashline{2-11}  \noalign{\smallskip}

IMC Dataset & R2D2-SS & 96.0 & 52k & 8.35 & \textbf{0.93} & R2D2-MS & 97.9 & 86k & 14.19 & \textbf{1.05} \\
(10 scenes) & w/ ScaleNet & \textbf{97.2} & \textbf{89k} & \textbf{10.71} & 1.02 & w/ ScaleNet & \textbf{99.0} & \textbf{114k} & \textbf{15.35} & 1.13 \\
 & w/ D-ScaleNet & 96.4 & 72k & 9.43 & 1.01 & w/ D-ScaleNet & 97.7 & 95k & 14.37 & 1.08 \\
\cdashline{2-11}  \noalign{\smallskip}

& Key.Net-SS & 99.1 & 82k & 11.05 & \textbf{1.06} & Key.Net-MS & 99.5 & 114k & 16.23 & \textbf{0.96} \\
& w/ ScaleNet & \textbf{99.6} & \textbf{148k} & \textbf{14.22} & 1.16 & w/ ScaleNet & \textbf{99.6} & \textbf{148k} & \textbf{19.23} & 1.09 \\
& w/ D-ScaleNet & 99.2 & 126k & 12.71 & 1.10 & w/ D-ScaleNet & 99.5 & 147k & 17.62 & 1.02 
\end{tabular}
\end{center}
\normalsize
\vspace{-0.30cm}
\caption{3D reconstruction results on the 10 test scenes (100 images each) from the IMC dataset \cite{imwb2020} with and without ScaleNet correction.}
\label{tab:3d_results}
\end{table*}

\noindent
\textbf{Protocol.} We also evaluate ScaleNet on the geometric correspondence task by integrating it into the popular  DGC-Net \cite{Melekhov2018} and GLU-Net \cite{GLUNet_Truong_2020}. Note that ScaleNet can also be combined with other recent methods \cite{truong2021gocor, truong2021learning, sun2021loftr}. ScaleNet rescales one of the images before the dense correspondence network estimates the dense flow fields between the two images. Due to the lack of dense annotations on real image pairs with large viewpoint and illumination changes, we evaluate ScaleNet using sparse correspondences available in the Megadepth \cite{li2018megadepth} dataset. Specifically, we follow the protocol of 1,600 image pairs introduced in \cite{shen2020ransac} to compute the percentage of correct keypoints (PCK) under a 5 pixel acceptance threshold. The experiment is extended with multiple acceptance thresholds in appendix \ref{appendix_sec:experiments_dense}.\\

\noindent
\textbf{Results} in table \ref{tab:dense_matching_megadepth_v2} show the PCK scores obtained with and without scale correction before the dense architectures, DGC-Net and GLU-Net. Moreover, to highlight the benefit of ScaleNet for images at different scale factors, besides reporting the results for the full dataset (\textit{All}), we create the \textit{Easy} ($s>1.2$) and \textit{Hard} ($s>1.8$) splits, where $s$ indicates scale distortions factor between images. Each split has 1,600, 627, and 440 image pairs, respectively. We observe that ScaleNet integration does not bring significant improvements  in the \textit{All} data, where the majority of images have small scale variations, but it is important to note that  it does not hurt the performance either. Meanwhile, in extreme cases (\textit{Hard} split),  current dense correspondence methods fail under severe scale changes, and their integration with ScaleNet improves the results by 347\% and 148\% for DGC-Net and GLU-Net.

\subsection{3D reconstruction}
\label{sec:3d_results}
\noindent
\textbf{Protocol.} ScaleNet can be easily integrated into geometric correspondence or relative camera pose pipelines, which are often a part of a more general 3D reconstruction system. To evaluate ScaleNet in this scenario, we follow the protocol proposed in the Local Feature Evaluation Benchmark \cite{schonberger2017comparative} for building 3D reconstruction models. 
As ScaleNet can upsample one of the images, and that could result in a higher number of candidate keypoints, we limit the number of features to the top 2,048 keypoints based on the protocol proposed in \cite{imwb2020}.
We present results for the test split from the IMC dataset \cite{imwb2020}, which includes ten different scenes with 100 images each. IMC images pose significant challenges, \textit{e.g.}, weather/illumination, perspective, scale, as well as strong occlusions.\\
\indent
As ScaleNet is applied to image pairs before the feature extraction, the detectors and descriptors are recomputed  every time the scale is corrected. Hence, to reduce the computation time, we propose a discrete variant of ScaleNet (\mbox{D-ScaleNet}) that makes the extraction process more efficient. \mbox{D-ScaleNet} implements a hard-assignment by selecting the maximum scale instead of the soft-scale and consistency check from equation \ref{eq:scale} and \ref{eq:cyclic}. Analogous to multi-scale pyramid approaches, we run the detectors/descriptors at multiple resized images but then select the optimal set of pre-computed features for matching based on D-ScaleNet estimation.\\

\begin{table}[]
\footnotesize
\begin{center}
\hspace{-0.25cm}
\begin{tabular}{c c c c c }
\multicolumn{1}{c}{}&\multicolumn{4}{c}{\textbf{Time (s)}}\\
\cline{2-5} \noalign{\smallskip}
& Extraction & Matching & Reconst. & Total\\
\hline \noalign{\smallskip}
SuperPoint & 10.3 & 16.7 & 195.3 & 222.3 \\
w/ ScaleNet & 980.5 & 46.3 & 208.9 & 1235.7 \\
w/ D-ScaleNet & 141.8 & 161.8 & 205.6 & 509.2
\end{tabular}
\end{center}
\normalsize
\vspace{-0.30cm}
\caption{3D reconstruction times on the \textit{British Museum} scene (100 images) from IMC dataset \cite{imwb2020}.}
\label{tab:efficiency_3d}
\vspace{-0.20cm}
\end{table}
\noindent
\textbf{Results} in table \ref{tab:3d_results} show the 3D reconstruction metrics of state-of-the-art methods with and without ScaleNet. 
We notice that image rectification by both, D-ScaleNet and ScaleNet, increases the number of registered images and the total number of observations in the 3D models. Track length is especially boosted by scale correction, meaning that the model was able to match the same keypoint simultaneously in more images. This increase of track length is particularly important since it proves that ScaleNet helps current methods distinguish and link points that  were not possible without it, due to extreme view differences. On average, improvements added by ScaleNet are greater than those produced by D-ScaleNet, however, D-ScaleNet still brings a notable boost over baselines. On the opposite side, ScaleNet increases the reprojection error (Rep. Error) of the reconstructions by 0.09 points on average. We attribute this to their longer track lengths since more points are triangulated throughout the images, and thus, the reprojection error increases, which has also been reported in \cite{schonberger2017comparative}. Longer tracks will benefit works that rely on complete and long tracks to refine the point positions and reduce their reprojection errors \cite{lindenberger2021pixel,Dusmanu2020Multi}. In table \ref{tab:efficiency_3d}, we show a comparison of the times taken to generate a 3D model when using ScaleNet and \mbox{D-ScaleNet}, and display the benefits in terms of computational time that D-ScaleNet provides.

\subsection{Image matching}
\begin{figure}[]
\hspace{-0.35cm}
 \centering
  \includegraphics[scale=0.32]{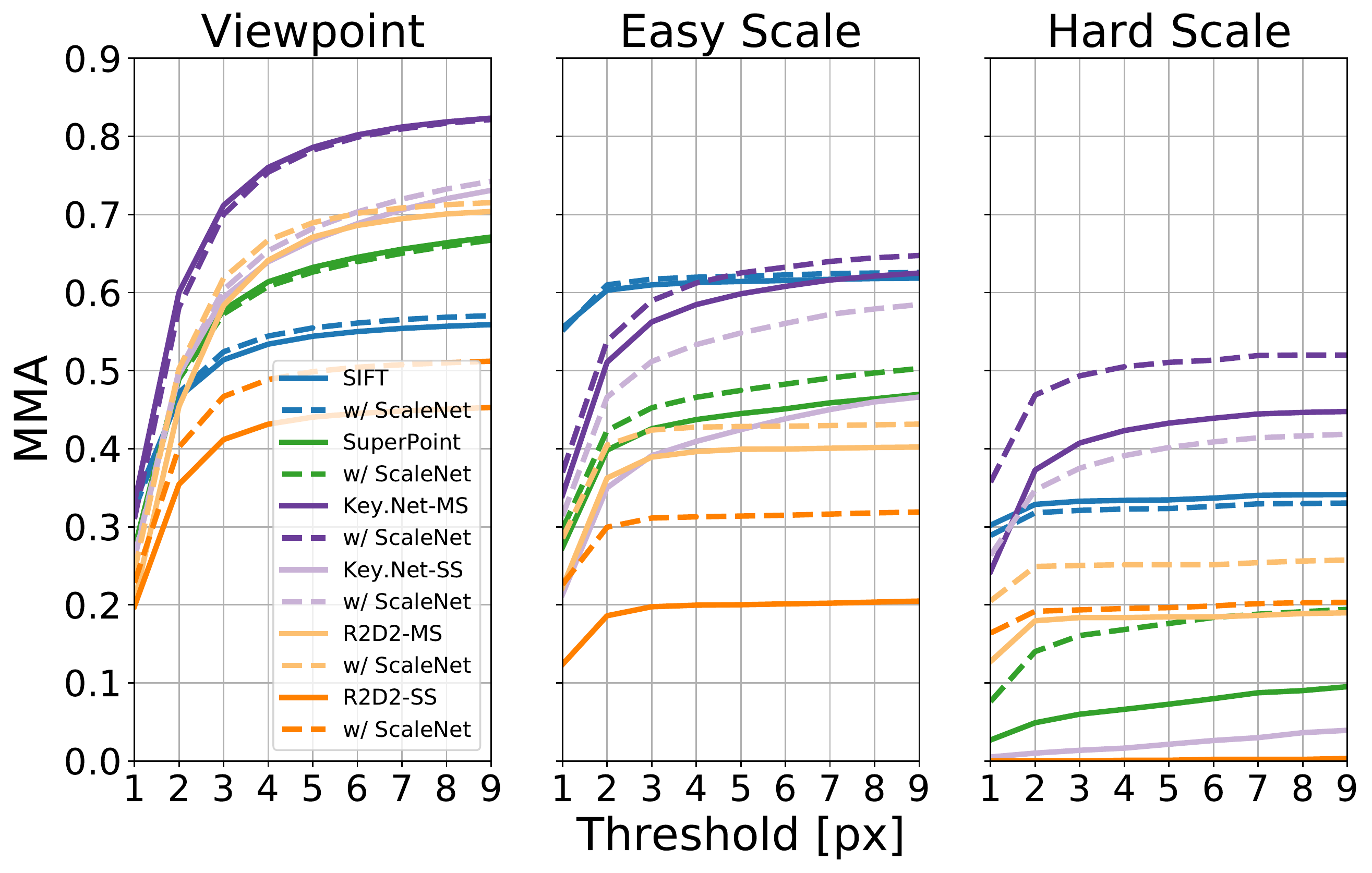}
    \caption{MMA on the full Viewpoint HPatches (left), scenes with easy (middle) and hard (right) scale transformations. Dashed lines \mbox{(- -)} show ScaleNet combined with other methods.}
    \label{fig:HPatches}
\end{figure}
\label{sec:matching}
\noindent
\textbf{Protocol.} We compute the Mean Matching Accuracy (MMA) \cite{mikolajczyk2005performance} as the ratio of correctly matched features within a threshold (5 pixels) and the total number of features following the benchmark proposed in \cite{dusmanu2019d2}. We report results for the 59 viewpoint scenes from the HPatches dataset \cite{balntas2017hpatches}. Similar to the 3D task, we fix the number of features to top 1,000 as in \cite{Karel_Vedaldi_BMVC_18} to eliminate the effect of increased  matching scores from a high number of keypoints. \\

\noindent
\textbf{Results} in figure \ref{fig:HPatches} (left) show that ScaleNet improves the robustness of current methods overall, and excels for single-scale methods, \textit{e.g.}, Key.Net-SS \cite{laguna2019key}, or \mbox{R2D2-SS \cite{revaud2019r2d2}}. However, in contrast to images used for statistics in figure \ref{fig:ablation_global}, most of the scenes in the HPatches Viewpoint contain strong perspective changes, \textit{i.e.}, display different scale factors within the scene. Hence, to highlight the effect of ScaleNet, we show results for the subset of scenes with affine transformations, \textit{i.e.}, scenes with stretch and skew transformations in addition to global scale and rotation. We select the splits such that the \textit{Easy} has $s \in [1.2,\ldots,1.8]$ and the \textit{Hard} has $s > 1.8$. When the scale factor is global across the image, ScaleNet correction can deal with planar scenes and improve the matching accuracy of single and multi-scale methods. As expected, results show that ScaleNet scaling is more critical for stronger scale changes in figure \ref{fig:HPatches} (right). Only \mbox{SIFT \cite{DoG}}, which is specially designed to be robust against scale changes on planar scenes, does not benefit from ScaleNet correction in the \textit{Hard} split. Moreover, to deal with scenes with possible strong perspective scale changes, we introduce in appendix \ref{appendix_sec:architecture_details_local} \mbox{Local-ScaleNet}, which infers local scale factors and offers a more robust and functional alternative for scenes with perspective changes.

\section{Discussion}
\noindent
\textbf{Limitations.} ScaleNet deals with arbitrary scale changes and, hence, it only brings improvements if such changes are present in the images. This makes ScaleNet useful in applications where those viewpoint changes prevent a successful matching of images, \textit{e.g.}, extreme and sparse collection of images, or ground-aerial applications. Nevertheless, even though ScaleNet does not boost performance when there is no scale change, it does not hurt either (cf. appendix \ref{appendix_sec:experiments_camera}). Another limitation comes when ScaleNet needs to be applied to a large collection of images, \textit{e.g.}, 3D tasks. ScaleNet works with pairs of images, hence, features can not be stored but need to be computed every time a new image is presented, increasing the feature extraction time as seen in table \ref{tab:efficiency_3d}. Although we propose D-ScaleNet, which mitigates the complexity time for such tasks, ScaleNet can be further optimized for faster processing by replacing VGG with more compact models such as MobileNet \cite{MobileNets_CKWWAA17}, or by only using ScaleNet for computing camera pose after a
restrictive retrieval search.\\

\noindent
\textbf{Societal impact.} Image matching is a pivotal but small component within large systems that facilitate technologies like AR, 3D reconstruction, navigation, modeling, SLAM, among others. Hence, as we contribute towards more robust matching pipelines, ScaleNet's societal impact is tied to the applications that rely on such technologies. Some applications may include smartphone apps, AR headsets, or autonomous cars. However, as our method cannot work independently of a larger system, the negative or ethical issues are not directly associated with our approach but rather with the specific business and final application where image matching may be used.\\

\noindent
\textbf{Reproducibility.} The experiments are computed on standard and public datasets and tasks, and hence, they can be reproduced. Moreover, we made public the evaluation and training scripts, as well as our custom training dataset. In addition, to encourage the research on scale estimation, we published the test set of section \ref{sec:camera_pose_megadepth} and splits of section \ref{sec:geometric_matching} for easier comparison and support of future works.

\section{Conclusions}

We introduced ScaleNet, an approach that estimates the scale change between images and improves the performance of methods that search for correspondences throughout different views of the same scene.
We proposed a novel learning scheme that formulates the problem of scale estimation as a prediction of a probability distribution of scales. We demonstrated how to make use of images from \mbox{non-planar} scenes to generate the training data. In addition to ScaleNet, we also introduced \mbox{D-ScaleNet}, a discrete variant of the proposed approach, and demonstrated its effectiveness in 3D-related tasks as well as computational time. We proved that ScaleNet can improve the results of popular pipelines in image matching for relative camera pose or 3D reconstruction while not being limited only to these tasks. 

\textbf{Acknowledgements.} This project was supported by Chist-Era EPSRC IPALM EP/S032398/1 grant.

\begin{appendices}

\section{Dataset details}
\label{appendix:implementation}

\begin{figure}[]
 \centering
  \includegraphics[scale=0.41]{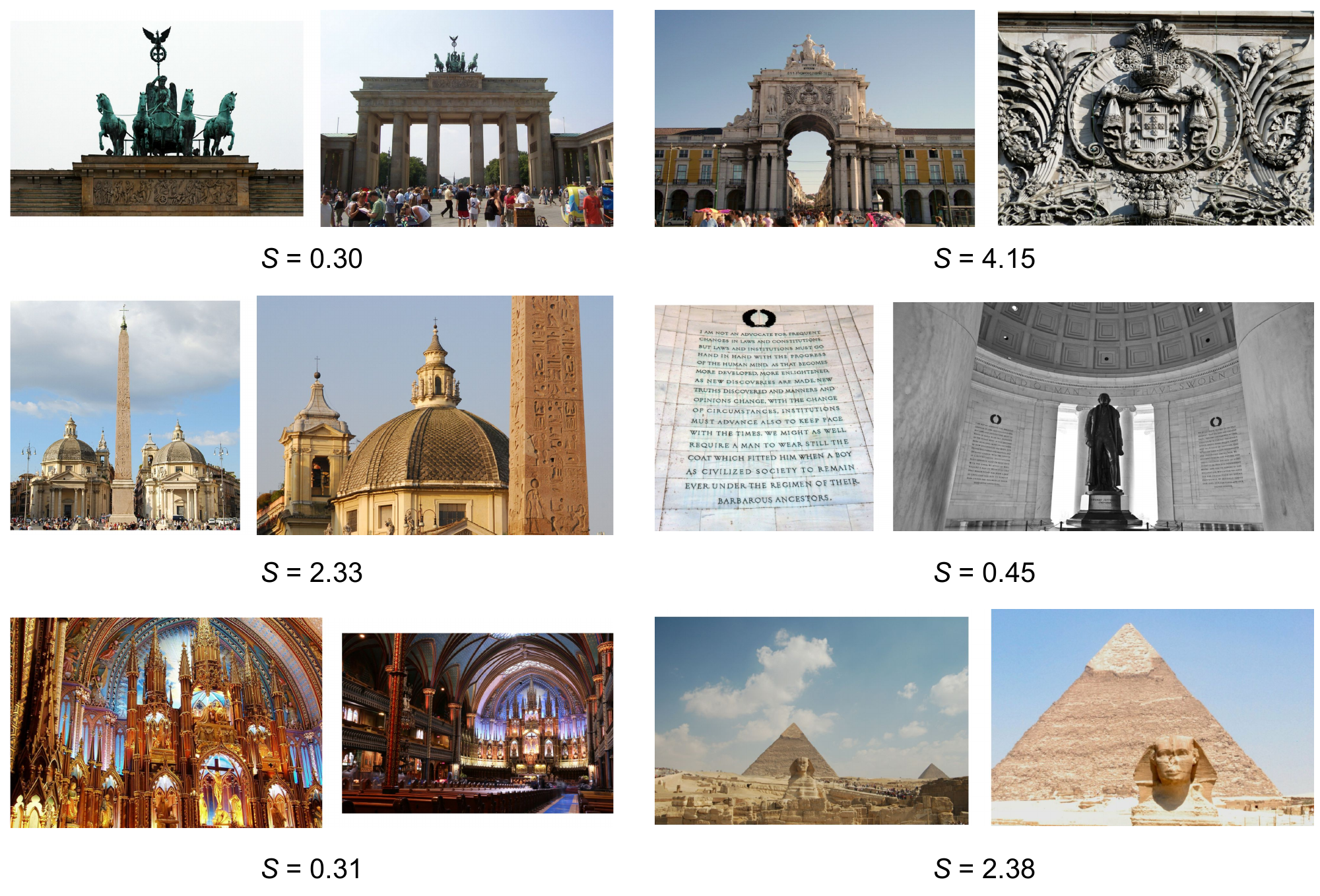}
    \caption{Image pairs examples of our dataset with their scale factor annotations computed as the average on logarithmic space of local scale changes.}
    \label{appendix_fig:im_examples}
\end{figure}

Real pair images are obtained directly from the Megadepth dataset \cite{li2018megadepth} as detailed in the main paper. As a reference, we display in figure \ref{appendix_fig:im_examples} examples of our dataset with their labeled scale changes. Since our labeling strategy relies on a 3D model, we can capture extreme scale changes between images if the collection of images that was used for the 3D reconstruction was big. While keypoint/descriptors may not be able to match correctly two images when there are strong scale changes, we rely on the fact that 3D models can relate two images if there were images in-between, \textit{i.e.}, we can compute the relationship between image $A$ and $B$ if we have an easier-to-match image $C$ in between the two views. This idea is somehow similar to the strategy proposed in MODS \cite{mishkin2015mods}, where authors create synthetic images between two different views to be able to relate them. In addition, we also use synthetic pairs to train ScaleNet. Synthetic pairs are computed on the fly, and therefore an unlimited number of examples can be created. We define a set of synthetic transformations with affine parameters: \mbox{scale $\in [0.16, 6]$}, \mbox{rotation $\in [-30$\textdegree$, 30$\textdegree$]$} and \mbox{skew $\in [-0.2, 0.2]$}. We use the random transformation to wrap the input image and generate the training pair. The ground-truth scale is directly the scale factor used to transform the image. \\


\section{Local-ScaleNet}
\label{appendix_sec:architecture_details_local}
Local-ScaleNet predicts a dense scale factor map between a pair of input images. In contrast to ScaleNet, in which scale estimation is global, pixel-wise \mbox{Local-ScaleNet} estimations offer a more suitable option when scale transformation significantly varies across the images, \textit{e.g.}, images with strong perspective distortion. \\
\begin{figure*}[]
 \centering
  \includegraphics[scale=0.80]{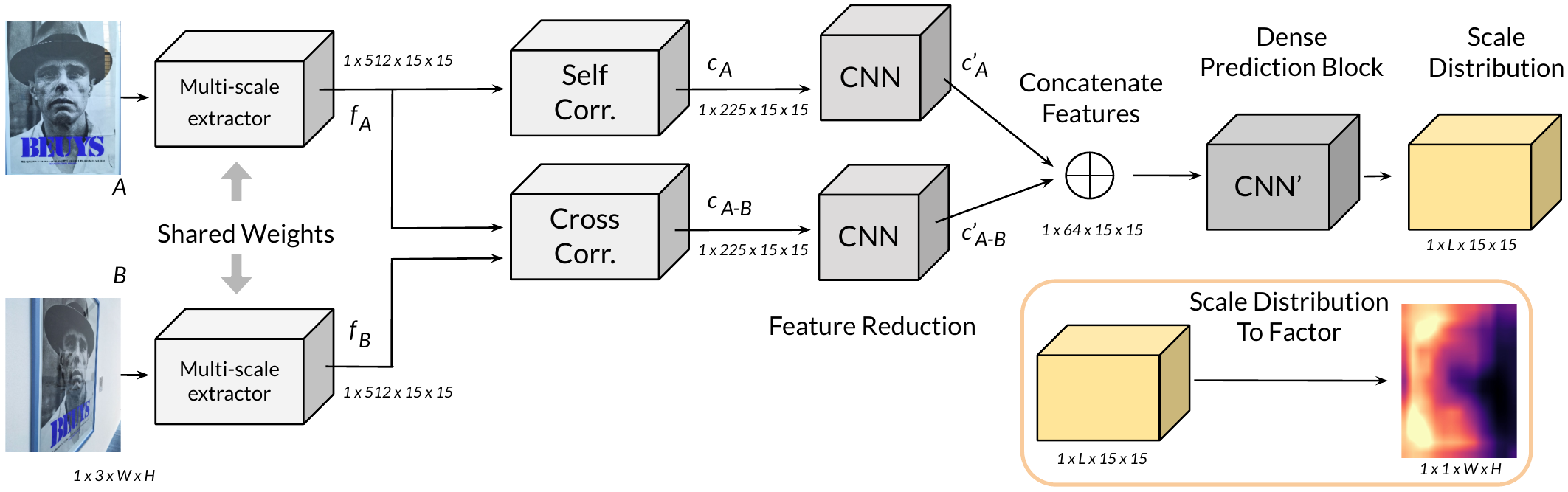}
    \caption{Local-ScaleNet uses a pre-trained VGG-16 and an ASPP block as its multi-scale feature extractor. After features are computed, a combination of self- and cross-correlation layers are used to calculate the relationship within image $A$ and between images $A$ and $B$. Correlation volumes' dimensionality is reduced through a CNN, and its results are concatenated into a single map. Finally, a dense scale distribution map is calculated by the final dense prediction block. During inference time, the scale distribution map is converted into a scale factor map and resized to the original resolution of image $A$.}
    \label{appendix_fig:local_architecture}
\end{figure*}

\noindent
\textbf{Architecture}. Local-ScaleNet architecture follows the scheme proposed in ScaleNet (cf. section \ref{sec:method}) with small variations. Analogous to ScaleNet, Local-ScaleNet takes two input images, $A$ and $B$, and computes features, $f_A$ and $f_B$, with the same multi-scale feature extractor block. Features, $f_A$ and $f_B$, go into the correlation layers to calculate the self- and cross-similarities, $c_A$ and $c_{A-B}$. In contrast to ScaleNet, Local-ScaleNet does not compute the self-similarities in image $B$. Since Local-ScaleNet estimates the scale factor locally, self-similarities in image $B$, $c_B$, are not spatially correlated with image $A$, and therefore, $c_B$ is not used in the per-pixel scale estimation. Similar to ScaleNet, $c_{A-B}$ and $c_A$ go into our local feature reduction block to process and reduce the channel dimension of the correlation maps. $c'_A$ and $c'_{A-B}$ are then concatenated and fed into the final dense prediction network which outputs the local distribution of scales between images $A$ and $B$. Furthermore, during test time, we transform our scale distribution into a scale factor map by following the pipeline presented in the main paper (cf. equation \ref{eq:cyclic}) in a pixel-wise manner. Due to image $A$ and $B$ not correlating spatially, we do not apply the consistency check from the original ScaleNet. Finally, the scale factor map is resized into the original image $A$ resolution. Full Local-ScaleNet architecture and pipeline are shown in figure \ref{appendix_fig:local_architecture}.\\

\begin{figure}[h!]
 \centering
  \includegraphics[scale=0.45]{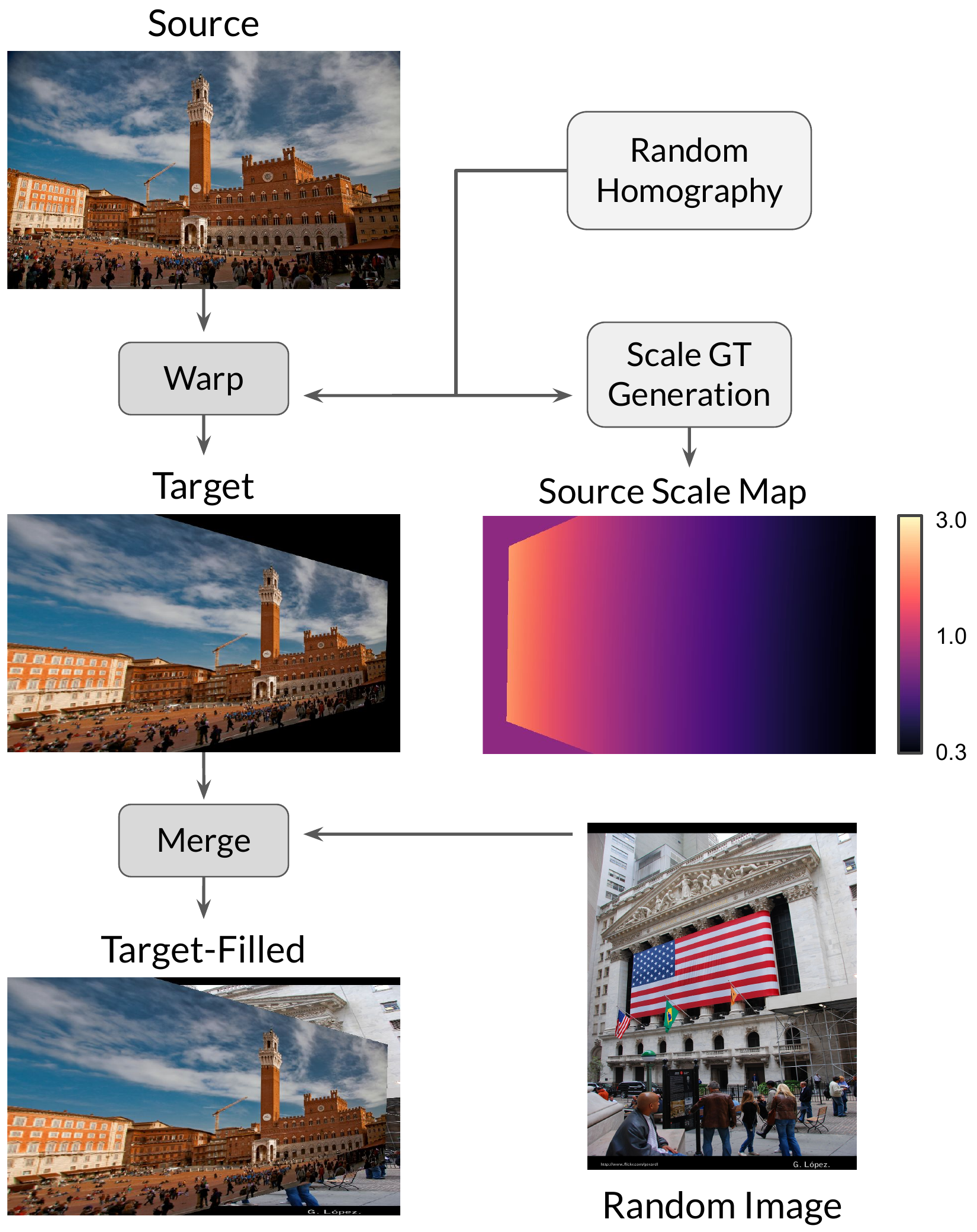}
    \caption{The local dataset generation pipeline, firstly, computes a random homography transformation. The homography is applied to an input image to generate a tuple of images. We refer to them as \textit{Source} and \textit{Target} images. The same homography is used to calculate the source scale map. Moreover, we take an extra random image from the dataset to fill those regions with zero values in the \textit{Target} image due to the homography transformation.}
    \label{appendix_fig:local_dataset}
\end{figure}
\noindent
\textbf{Dataset.} As discussed, Local-ScaleNet generates a dense map of scale distributions. To train it, we need pairs of images with their corresponding pixel-wise scale ground-truths. We based our dataset on synthetic pairs of images as displayed in figure \ref{appendix_fig:local_dataset}. Even though real pairs of images could be used to train Local-ScaleNet, real image scenes do not present strong local distortions, \textit{i.e.}, global scale often reflects the scale changes in local regions. Thus, we use synthetic pairs to ensure that the global scale differs from the scales between local regions. \\
\indent
A random homography transformation is generated and applied to an input image, such that the input image, \textit{Source}, and the warped one, \textit{Target}, are geometrically related by our homography. We use the same parameters to build the homography matrix as those presented in appendix \ref{appendix:implementation} and add a perspective distortion to ensure different local scale factors throughout the \textit{Target} image. Moreover, due to strong homography transformations, \textit{Target} image may contain zero values in the non-overlapping regions as seen in figure \ref{appendix_fig:local_dataset}. To avoid the network being driven by those zero values, we filled the non-overlapping regions between \textit{Source} and \textit{Target} images with a randomly sampled image from the dataset. Therefore, our final training pair is composed by the \textit{Source} and our new \textit{Target-Filled} image. Lastly, we use the same homography to generate the ground-truth scale map between \textit{Source} and \textit{Target-Filled}. As we have the synthetic transformation, we generate the source scale map by sampling local points throughout the whole \textit{Source} image and computing the scale factor within their neighborhood region as the ratio of their distances in the \textit{Source} and \textit{Target-Filled} images.

\section{Extended experiments}
\label{appendix_sec:experiments}
This section extends the experiments and results presented in the main paper. 

\subsection{Design choices}
\label{appendix_sec:experiments_ablation}
ScaleNet embraces some key ideas on its learning scheme and parameterization that help towards delivering good scale estimations. We discuss them here and highlight the importance of our decision during ScaleNet design. Table \ref{appendix_tab:ablation_study} shows the performance of our baseline, which uses SuperPoint network without any scale rectification. Moreover, we report the results of ScaleNet without the consistency check for an easier comparison against other designs. \\

\noindent
\textbf{Natural vs logarithmic space}. The scale factor is a relative ratio operator, hence, it is non-linear. We compute the scale factor as a soft-computation based on quantized scale classes, $s_i$, and a probability distribution (cf. equation \ref{eq:scale}). To avoid being biased by high scale values when computing the soft-scale, we transform the quantized scale classes, $s_i$ , to logarithmic space. Hence, to demonstrate the superiority of this parameterization, we display in table \ref{appendix_tab:ablation_study} the results of doing the soft-computation directly in natural space (\textit{Natural rep.}). We see that even though the performance of the natural representation model is over the baseline, we can improve upon it by the simple and effective approach of dealing with the scale factors in the logarithmic space.\\

\noindent
\textbf{Regression vs classification model}. Even though predicting directly the scale factor between images is theoretically possible, in the practical scenario, it appears to be a much harder task. To prove it, we have trained a regression model with the L2 loss, where our network had to predict a normalized scale factor between the two images in logarithmic space. We use logarithmic space since the previous experiment shows the benefits of this representation. As shown in table \ref{appendix_tab:ablation_study}, the proposed ScaleNet regressor model does not bring any benefit over the baseline accuracy (which does not use any scale rectification). We claim that even though a regressor model could be trained, learning and interpreting the relationships between the quantized scale ranges is an easier task, and hence, we embrace the decision of a classification model rather than a regressor.\\
\begin{table}[]
\footnotesize
\begin{center}
\begin{tabular}{c c c c}
\multicolumn{1}{c}{} & \multicolumn{3}{c}{\textbf{Pose estimation (AUC)}}\\ 
\cline{2-4} 
\noalign{\smallskip}
 & at 5\textdegree & at 10\textdegree & at 20\textdegree \\
\hline \noalign{\smallskip}
Baseline & 4.8 & 7.4 & 10.4 \\
\hdashline\noalign{\smallskip}
ScaleNet & \textbf{8.4} & \textbf{12.3} & \textbf{17.6} \\
\hdashline\noalign{\smallskip}
Natural rep. & 7.2 & 11.2 & 15.9 \\
Regression & 4.9 & 7.3 & 10.1 \\
D-ScaleNet (hard-assig.) & 7.9 & 11.5 & 16.7 \\
\end{tabular}
\end{center}
\normalsize
\vspace{-0.30cm}
\caption{Ablation study of the different ScaleNet's design choices. Baseline refers to SuperPoint \cite{detone2018superpoint} without scale correction, and ScaleNet refers to the method introduced in the paper without consistency check (cf. equation 3 main paper).}
\label{appendix_tab:ablation_study}
\end{table}

\noindent
\textbf{Hard-assignment vs soft-assignment}. Hard-assignment has been already introduced in the main paper as discrete ScaleNet (\mbox{D-ScaleNet}). \mbox{D-ScaleNet} uses the maximum predicted value of the scale distribution as the scale factor between images, instead of computing the soft-scale (cf. equation \ref{eq:scale}). Even though we provided the advantages and drawbacks in terms of computation time and 3D metrics (cf. section \ref{sec:3d_results}), we further display their differences in table \ref{appendix_tab:ablation_study}. Results show that the soft-assignment, and hence, being able to interpolate between the quantized scale values, provides better results than \mbox{D-ScaleNet}. However, \mbox{D-ScaleNet} can offer a faster alternative to ScaleNet and still bring notable improvements over the baseline.

\subsection{Image matching}
\begin{table}[]
\footnotesize
\begin{center}
\begin{tabular}{c c c c}
\multicolumn{4}{c}{\textbf{HPatches MMA-5px (\%)}}\\
\hline \noalign{\smallskip}
& Easy & Medium & Hard\\
\hline \noalign{\smallskip}
Key.Net-SS / HardNet & 66.65 & 47.40 & 25.52 \\
w/ ScaleNet & \textbf{71.98} & \textbf{48.40} & 21.42 \\
w/ Local-ScaleNet & 64.63 & 45.01 & \textbf{28.43}\\
\end{tabular}
\end{center}
\normalsize
\vspace{-0.30cm}
\caption{Mean matching accuracy (MMA) results on perspective sequences from HPatches \cite{balntas2017hpatches}.}
\label{appendix_tab:local_scale}
\end{table}
Local scale estimation allows to locally correct the scale of a given keypoint. Local estimations are effective when the scale transformation significantly varies across the image. Therefore, we test the effect of using the dense variant of ScaleNet (\mbox{Local-ScaleNet}) on sequences with a strong perspective change, \textit{i.e.}, with different local scale factors. Local-ScaleNet substitutes the fully connected layers of ScaleNet with the CNN dense prediction block detailed in section \ref{appendix_sec:architecture_details_local}. As the scale correction is local, we combine ScaleNet with Key.Net/HardNet \cite{laguna2019key,mishchuk2017working},  which allows correcting locally its feature extraction. Specifically, we run Key.Net in the original image and use the scale map estimated by \mbox{Local-ScaleNet} to correct the size of the extracted patch before computing a descriptor with HardNet. We split image pairs from HPatches \cite{balntas2017hpatches} into easy, medium, and hard subsets according to their perspective distortions. Table \ref{appendix_tab:local_scale} shows that using ScaleNet for local correction is effective for significant perspective changes but deteriorates the performance for less challenging transformations. Such behavior was also observed for scale and affine invariant handcrafted feature extractors \cite{mikolajczyk2005performance}.
Moreover, as \mbox{Local-ScaleNet} can only be applied to local descriptors, global ScaleNet is more suitable as a general approach.  

\subsection{IMC benchmark}

\begin{table*}[!ht]
\footnotesize
\begin{center}
\begin{tabular}{c c c c c c c c c }
\multicolumn{1}{c}{} & \multicolumn{8}{c}{mAA (\%) at 10\textdegree} \\
\cline{2-9} \noalign{\smallskip}

& \bf{Stereo} & \bf{Multiview} & \bf{Average} & & & \bf{Stereo} & \bf{Multiview} & \bf{Average} \\
\cline{2-5} \cline{7-9} \noalign{\smallskip}

SIFT & 43.1 & 43.2 & 43.2 & & SuperPoint & 32.9 & 54.5 & 43.7  \\
w/ ScaleNet & \textbf{43.4} & \textbf{47.3} & \textbf{45.4} & & w/ ScaleNet & \textbf{41.3} & \textbf{61.9} & \textbf{52.0}  \\
\cdashline{2-5} \cdashline{7-9}  \noalign{\smallskip}

R2D2-SS & 13.4 & 12.3 & 12.9 & & R2D2-MS & 34.9 & \textbf{48.0} & \textbf{41.5} \\
w/ ScaleNet & \textbf{26.3} & \textbf{31.7} & \textbf{29.0} & & w/ ScaleNet & \textbf{35.5} & 46.7  & 41.1 \\
\cdashline{2-5} \cdashline{7-9}  \noalign{\smallskip}

Key.Net-SS & 37.3 & 57.4 & 47.4 & & Key.Net-MS & 60.2 & 73.6 & 66.9\\
w/ ScaleNet & \textbf{56.1} & \textbf{71.9} & \textbf{64.0} & & w/ ScaleNet & \textbf{62.6} & 73.6 & \textbf{68.1} \\
\cline{1-9}  \noalign{\smallskip}

\end{tabular}
\end{center}
\normalsize
\caption{Mean Average Accuracy (mAA) at 10\textdegree on IMC dataset \cite{imwb2020}.}
\label{tab:camera_localization}
\end{table*}
\noindent
\textbf{Protocol.} We follow the protocol proposed in Image Matching Challenge \cite{imwb2020} for computing their two main tasks, wide-baseline stereo and multi-view reconstruction. The benchmark looks to the pose errors under different thresholds and reports the mean Average Accuracy (mAA) of the reconstructions. We follow the evaluation protocol with 2,048 keypoints. Moreover, we select DEGENSAC \cite{chum2005two} for geometric verification and first-to-second nearest-neighbour ratio for filtering false-positive matches. For more details on their evaluation refer to \cite{imwb2020}. We search for the best configuration in stereo and multi-view independently for each method. We use the validation scenes provided in Image Matching Competition\footnote{https://www.cs.ubc.ca/research/image-matching-challenge} as authors only made public the ground-truth for such scenes. Computing ScaleNet in the test set is not straightforward, since the benchmark had to be modified to accept a different set of features for each pair of images. Thus, as a reference to our method's effect in their benchmark, we report the best results obtained in the three validation scenes, \textit{Reichstag} (74 images), \textit{Sacre Coeur} (100 images), and \textit{St Peters Square} (100 images).\\

\noindent
\textbf{Results} in table \ref{tab:camera_localization} give the benefits of camera pose error when using ScaleNet. The boost of ScaleNet when working together with a single scale method shows the importance of rectifying the scale difference between images. Even though multi-scale methods look for features on multiple scaled images, the best results are obtained when ScaleNet corrects one of the images before feature extraction. Moreover, ScaleNet brings benefits to current pipelines when the scale change between image pairs is extreme, and therefore, if the scale factor is not significant as in the evaluated scenes, multi-scale feature extraction is sufficient to perform the correspondence search.

\subsection{Camera pose}
\begin{table}[]
\footnotesize
\begin{center}
\begin{tabular}{c c c c}
\multicolumn{1}{c}{} & \multicolumn{3}{c}{\textbf{Pose estimation (AUC)}}\\ 
\cline{2-4} 
\noalign{\smallskip}
 & at 5\textdegree & at 10\textdegree & at 20\textdegree \\
\hline \noalign{\smallskip}
SP + SuperGlue & 35.2 & 54.7 & 71.6 \\
w/ ScaleNet & \textbf{36.3} (+3\%) & \textbf{55.7} (+2\%) & \textbf{72.4} (+1\%) \\
\end{tabular}
\end{center}
\normalsize
\vspace{-0.30cm}
\caption{Camera pose results on Megadepth test split from \cite{sun2021loftr}. When scale changes are small on the image pairs, ScaleNet brings more robustness without hurting the performance.}
\label{appendix_tab:camera_pose_default_megadepth}
\end{table}
\begin{table*}[]
\scriptsize
\begin{center}
\begin{tabular}{c c c c | c c c | c c c}
\multicolumn{1}{c}{} & \multicolumn{3}{c}{\textbf{All}} & \multicolumn{3}{c}{\textbf{Easy}} & \multicolumn{3}{c}{\textbf{Hard}}\\
\hline \noalign{\smallskip}
& PCK-1 (\%) & PCK-5 (\%) & PCK-10 (\%)& PCK-1 (\%) & PCK-5 (\%) & PCK-10 (\%)& PCK-1 (\%) & PCK-5 (\%) & PCK-10 (\%)\\
\hline \noalign{\smallskip}
DGC-Net & 7.4 & 40.2 & 51.5 & 5.3 & 34.4 & 48.0 & 3.2 & 4.5 & 9.6\\
w/ ScaleNet & \bf{7.9} (+7\%) & \bf{41.5} (+3\%) & \bf{53.3} (+4\%) & \bf{5.9} (+11\%) & \bf{36.8} (+7\%) & \bf{51.4} (+7\%) & \bf{3.3} (+3\%) & \bf{20.1} (+347\%) & \bf{28.7} (+199\%)\\
\hdashline\noalign{\smallskip}
GLU-Net & 21.3 & 55.5 & 62.2 & 19.5 & 55.4 & 62.5 & 1.7 & 10.8 & 15.5\\
w/ ScaleNet & \bf{23.3} (+9\%) & \bf{57.8} (+4\%) & \bf{64.5} (+4\%) & \bf{20.4} (+5\%) & \bf{56.3} (+2\%) & \bf{64.5} (+3\%) & \bf{7.3} (+329\%) & \bf{26.8} (+148\%) & \bf{33.8} (+118\%)\\
\end{tabular}
\end{center}
\normalsize
\vspace{-0.30cm}
\caption{ Percentage of correct keypoints (PCK) for different pixel thresholds on MegaDepth \cite{li2018megadepth} sparse correspondences. The results are reported for DGC-Net \cite{Melekhov2018} and GLU-Net \cite{GLUNet_Truong_2020} without and with our ScaleNet, which consistently improves the performance.}
\label{appendix_tab:dense_matching}
\end{table*}
\label{appendix_sec:experiments_camera}
In addition to the results of the main paper, we extend in table \ref{appendix_tab:camera_pose_default_megadepth} the results of the camera pose task. We test our method together with state-of-the-art combination SuperPoint \cite{detone2018superpoint} and SuperGlue \cite{superglue2019} and report results in the standard Megadepth \cite{li2018megadepth} test split proposed in \cite{sun2021loftr}. This split contains 1,500 image pairs from two different scenes, where image pairs suffer from general viewpoint and illumination challenges but not extreme scale changes. Hence, we see in table \ref{appendix_tab:camera_pose_default_megadepth} that even though ScaleNet does not bring great improvements, it does not either hurt the performance when there are no strong scale changes between the image pairs. 

\subsection{Dense matching}
\label{appendix_sec:experiments_dense}
We also extend the evaluation on dense geometric matching methods, DGC-Net \cite{Melekhov2018} and GLU-Net \cite{GLUNet_Truong_2020}, by reporting the percentage of correct keypoints (PCK) under multiple pixel acceptance thresholds, \textit{i.e.}, 1, 5, and 10 pixels. We use the same sparse correspondence split proposed by \cite{shen2020ransac} in the Megadepth dataset \cite{li2018megadepth}. As in the main paper, we report the results for the full dataset (\textit{All}) and create the
\textit{Easy} ($s > 1.2$) and \textit{Hard} ($s > 1.8$) splits, where $s$ indicates scale distortions factor between images. Results in table \ref{appendix_tab:dense_matching} show that the improvements that ScaleNet brings into the dense methods are similar across all acceptance thresholds, offering extra robustness across all splits, and a large boost when the scale changes between images is strong (\textit{Hard} split).

\subsection{Evaluating scale predictions}
\begin{table}[]
\footnotesize
\begin{center}
\begin{tabular}{c c c c c}
\multicolumn{5}{c}{\textbf{Scale Accuracy ($r$)}}\\
\hline \noalign{\smallskip}
& Perfect & Random & Constant & ScaleNet\\
\hline \noalign{\smallskip}
 & 1.0 & 4.3 & 3.2 & \textbf{1.6} \\
\end{tabular}
\end{center}
\normalsize
\vspace{-0.30cm}
\caption{Scale prediction accuracy computed as the mean scale ratios, $r$, between the ground-truth, $S_{GT}$,  and estimated scales, $s$.}
\label{appendix_tab:eval_scale}
\end{table}
Despite the evaluation of ScaleNet in different 3D tasks, in this section, we introduce the results in scale prediction accuracy. Although the real impact of ScaleNet is measured in terms of 3D metrics, knowing its scale accuracy gives a better understanding of the capability of our scale predictor. \\

\noindent
\textbf{Protocol.} To test the accuracy of our scale predictions, we compute the mean scale ratio, $r$, between the ground-truth, $S_{GT}$, and the estimated scale, $s$. Specifically, we compute the ratio of a scale prediction as $r = s_a / s_b$, $s_a = max(S_{GT}, s)$, and $s_b = min(S_{GT}, s)$. In addition to ScaleNet predictions, we also propose two baselines as a reference: \textit{Random} makes a random prediction with $s \in [0.16, 6]$, and \textit{Constant} uses a fixed scale value such as $s=1.0$. In addition, we also indicate the results of perfect scale estimation in \textit{Perfect} ($r=1.0$).
We use the test set proposed in the relative camera pose experiment of section \ref{sec:camera_pose_megadepth}.\\

\noindent
\textbf{Results.} We report in table \ref{appendix_tab:eval_scale} the mean scale ratio between ground-truth and predicted scales. ScaleNet predictions ($r=1.6$) are accurate enough to bring images to scale discrepancies where current local feature extractors can operate. However, even though scale prediction accuracy is relevant, it is hard to quantify. Therefore, we believe that results on downstream applications are needed to understand the impact of ScaleNet image rectification.

\subsection{Complexity time}

\begin{table}[]
\footnotesize
\begin{center}
\begin{tabular}{c c c c c c }
\multicolumn{6}{c}{\textbf{Time (ms)}}\\
\hline \noalign{\smallskip}
& DGC-Net & GLU-Net & R2D2 & Key.Net & ScaleNet\\
\hline \noalign{\smallskip}
 & 35.4 & 25.1 & 384.3 & 640.5 & 19.8 \\
\end{tabular}
\end{center}
\normalsize
\vspace{-0.30cm}
\caption{Comparison of computational times in ms for several state-of-the-art methods. We report the multi-scale extraction times for R2D2 and Key.Net/HardNet. }
\label{appendix_tab:time_dense}
\end{table}
Table \ref{appendix_tab:time_dense} introduces the computational time of ScaleNet and other state-of-the-art methods. 
Although ScaleNet has the fastest inference time, it only predicts a single transformation parameter rather than the flow between images or local keypoints/descriptors. However, we showed that ScaleNet largely benefits other methods if scale changes are presented in the images and increases only 19.8ms the computational time of matching two images. Besides dense geometric matching models, \mbox{DGC-Net} (35.4ms) and \mbox{GLU-Net} (25.1ms), we also provided the times of multi-scale feature extraction of \mbox{R2D2-MS} (384.3ms) and \mbox{Key.Net-MS/HardNet} (640.5ms). As multi-scale methods require a minimum image size to work properly, we report the times when extracting features from an image of size \mbox{$600\times800$}, and see that the time for  ScaleNet is significantly smaller than the feature extractors. Thus, we believe that if scale robustness is needed, the improvements outweigh the computational overhead introduced by our scale correction.

Besides, some strategies could reduce the impact of using scale correction. As discussed in the main paper, ScaleNet could use faster and more efficient feature extractors, \textit{i.e.}, MobileNet \cite{MobileNets_CKWWAA17}, or we could precompute ScaleNet backbone features and store them.
Catching ScaleNet backbone features would allow us only to run correlation and fully connected layers for every new pairs of images, reducing the complexity of some 3D system, \textit{e.g.}, 3D reconstruction. As a reference, the time on inference when caching backbone features is 1.8ms, meanwhile, the inference time of regular ScaleNet is 19.8ms.

\section{Qualitative examples}
\label{appendix_sec:qualitative}
\begin{figure*}[th!]
 \centering
  \includegraphics[scale=0.135]{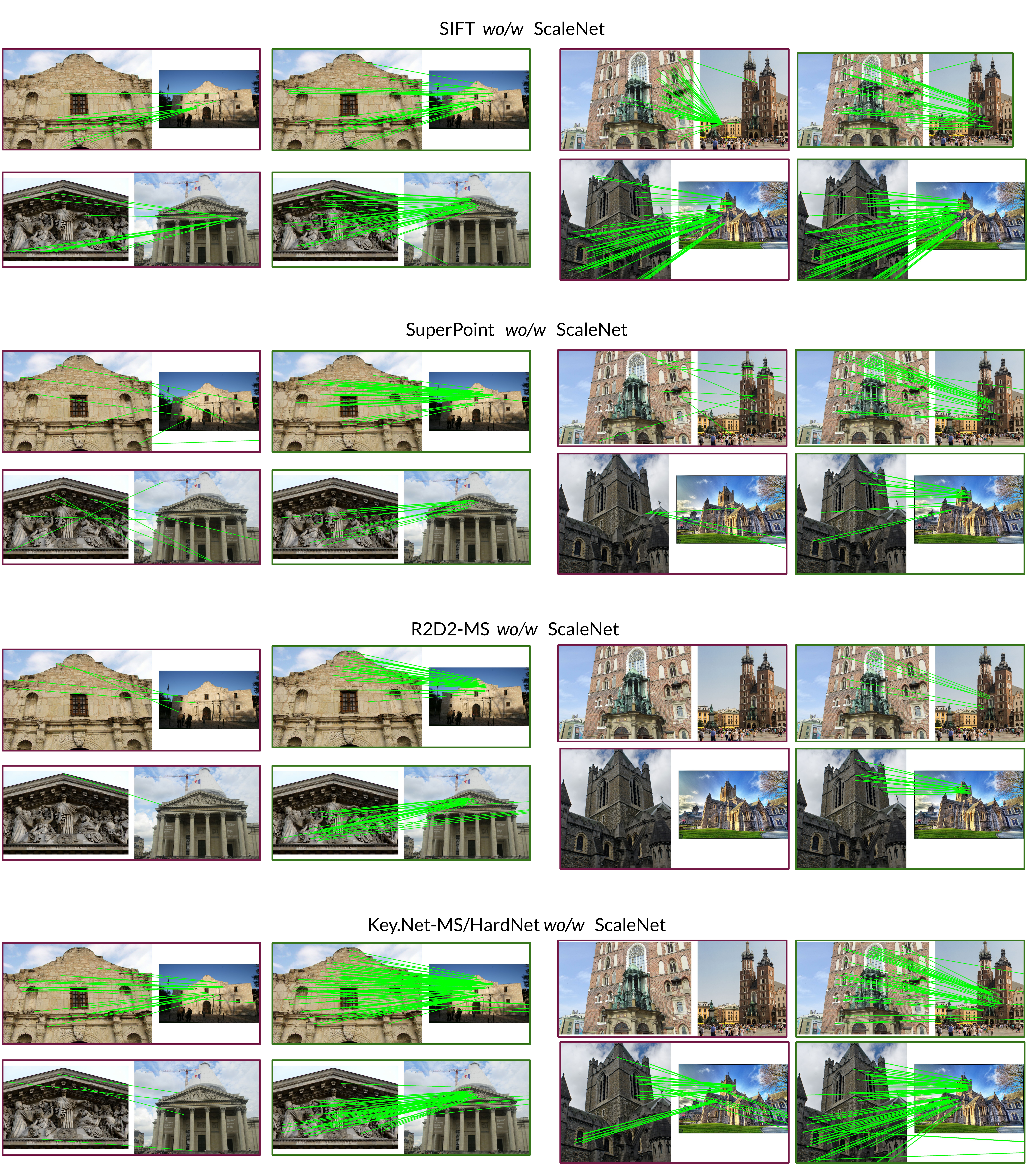}
    \vspace{0.30cm}
    \caption{Comparison of matches \textbf{without} (red boxes) and \textbf{with} (green boxes) the scale rectification proposed by ScaleNet. We only plot the matches that agree with the camera pose computed by the methods, either with or without ScaleNet. The examples display extreme scale conditions, and hence, we observe that the number of matches notably increases when applying ScaleNet's image rectification.}
    \label{appendix_fig:qualitative}
\end{figure*}

In this section, we provide some qualitative results from applying ScaleNet in image pairs from the Megadepth dataset \cite{li2018megadepth} that contain strong scale changes. In figure \ref{appendix_fig:qualitative}, we show the matches that were obtained for each method. We only plot the matches that survive the Lowe’s ratio test\cite{DoG} and MAGSAC\cite{barath2019magsac} geometric fitting method. As in the paper, to avoid the effect of a higher number of keypoints due to one image being upsampled, we only use the top 2,048 keypoint candidates. We observe that all methods benefit from the scale correction when images present severe scale variations. Furthermore, examples show difficult image pairs where the multi-scale methods, SIFT \cite{DoG}, R2D2 \cite{revaud2019r2d2}, and Key.Net/HardNet \cite{laguna2019key, mishchuk2017working}, could not handle the extreme-scale change and found few or non-matches between pairs. We see that on those examples, methods benefit largely from scale rectification. Similarly, SuperPoint \cite{detone2018superpoint} could only find correct matches if the scale factor between images is corrected.

\end{appendices}

{\small
\bibliographystyle{ieee_fullname}
\bibliography{egbib}

\begin{thebibliography}{10}\itemsep=-1pt

\bibitem{balntas2017hpatches}
Vassileios Balntas, Karel Lenc, Andrea Vedaldi, and Krystian Mikolajczyk.
\newblock Hpatches: A benchmark and evaluation of handcrafted and learned local
  descriptors.
\newblock In {\em Proceedings of the IEEE Conference on Computer Vision and
  Pattern Recognition}, pages 5173--5182, 2017.

\bibitem{barath2019magsac}
Daniel Barath, Jiri Matas, and Jana Noskova.
\newblock Magsac: marginalizing sample consensus.
\newblock In {\em Proceedings of the IEEE/CVF Conference on Computer Vision and
  Pattern Recognition}, pages 10197--10205, 2019.

\bibitem{laguna2019key}
Axel Barroso-Laguna, Edgar Riba, Daniel Ponsa, and Krystian Mikolajczyk.
\newblock Key.net: Keypoint detection by handcrafted and learned cnn filters.
\newblock In {\em Proceedings of the IEEE International Conference on Computer
  Vision}, 2019.

\bibitem{hdd2020}
Axel Barroso-Laguna, Yannick Verdie, Benjamin Busam, and Krystian Mikolajczyk.
\newblock Hdd-net: Hybrid detector descriptor with mutual interactive learning.
\newblock In {\em Proceedings of the Asian Conference on Computer Vision},
  2020.

\bibitem{brachmann2019neural}
Eric Brachmann and Carsten Rother.
\newblock Neural-guided ransac: Learning where to sample model hypotheses.
\newblock In {\em Proceedings of the IEEE/CVF International Conference on
  Computer Vision}, pages 4322--4331, 2019.

\bibitem{burki2019vizard}
Mathias B{\"u}rki, Lukas Schaupp, Marcin Dymczyk, Renaud Dub{\'e}, Cesar
  Cadena, Roland Siegwart, and Juan Nieto.
\newblock Vizard: Reliable visual localization for autonomous vehicles in urban
  outdoor environments.
\newblock In {\em 2019 IEEE Intelligent Vehicles Symposium (IV)}, pages
  1124--1130. IEEE, 2019.

\bibitem{chen2017deeplab}
Liang-Chieh Chen, George Papandreou, Iasonas Kokkinos, Kevin Murphy, and Alan~L
  Yuille.
\newblock Deeplab: Semantic image segmentation with deep convolutional nets,
  atrous convolution, and fully connected crfs.
\newblock {\em IEEE transactions on pattern analysis and machine intelligence},
  40(4):834--848, 2017.

\bibitem{chen2017rethinking}
Liang-Chieh Chen, George Papandreou, Florian Schroff, and Hartwig Adam.
\newblock Rethinking atrous convolution for semantic image segmentation.
\newblock {\em arXiv preprint arXiv:1706.05587}, 2017.

\bibitem{chum2005two}
Ondrej Chum, Tomas Werner, and Jiri Matas.
\newblock Two-view geometry estimation unaffected by a dominant plane.
\newblock In {\em Conference on Computer Vision and Pattern Recognition},
  volume~1, pages 772--779. IEEE, 2005.

\bibitem{imagenet_cvpr09}
J. Deng, W. Dong, R. Socher, L.-J. Li, K. Li, and L. Fei-Fei.
\newblock {ImageNet: A Large-Scale Hierarchical Image Database}.
\newblock In {\em Proceedings of the IEEE Conference on Computer Vision and
  Pattern Recognition (CVPR)}, 2009.

\bibitem{detone2018superpoint}
Daniel DeTone, Tomasz Malisiewicz, and Andrew Rabinovich.
\newblock Superpoint: Self-supervised interest point detection and description.
\newblock In {\em Proceedings of the IEEE Conference on Computer Vision and
  Pattern Recognition Workshops}, pages 224--236, 2018.

\bibitem{dusmanu2019d2}
Mihai Dusmanu, Ignacio Rocco, Tomas Pajdla, Marc Pollefeys, Josef Sivic,
  Akihiko Torii, and Torsten Sattler.
\newblock D2-net: A trainable cnn for joint detection and description of local
  features.
\newblock In {\em Proceedings of the IEEE Conference on Computer Vision and
  Pattern Recognition}, 2019.

\bibitem{Dusmanu2020Multi}
Mihai Dusmanu, Johannes~L. Sch\"onberger, and Marc Pollefeys.
\newblock {Multi-View Optimization of Local Feature Geometry}.
\newblock In {\em Proceedings of the European Conference on Computer Vision},
  2020.

\bibitem{ebel2019beyond}
Patrick Ebel, Anastasiia Mishchuk, Kwang~Moo Yi, Pascal Fua, and Eduard Trulls.
\newblock Beyond cartesian representations for local descriptors.
\newblock In {\em Proceedings of the IEEE International Conference on Computer
  Vision}, pages 253--262, 2019.

\bibitem{germain2020s2dnet}
Hugo Germain, Guillaume Bourmaud, and Vincent Lepetit.
\newblock S2dnet: Learning accurate correspondences for sparse-to-dense feature
  matching.
\newblock In {\em European Conference on Computer Vision}, 2020.

\bibitem{he2016deep}
Kaiming He, Xiangyu Zhang, Shaoqing Ren, and Jian Sun.
\newblock Deep residual learning for image recognition.
\newblock In {\em Proceedings of the IEEE conference on computer vision and
  pattern recognition}, pages 770--778, 2016.

\bibitem{MobileNets_CKWWAA17}
Andrew~G. Howard, Menglong Zhu, Bo Chen, Dmitry Kalenichenko, Weijun Wang,
  Tobias Weyand, Marco Andreetto, and Hartwig Adam.
\newblock Mobilenets: Efficient convolutional neural networks for mobile vision
  applications.
\newblock {\em CoRR}, abs/1704.04861, 2017.

\bibitem{jaderberg2015spatial}
Max Jaderberg, Karen Simonyan, Andrew Zisserman, et~al.
\newblock Spatial transformer networks.
\newblock In {\em Advances in neural information processing systems}, pages
  2017--2025, 2015.

\bibitem{jiang2021cotr}
Wei Jiang, Eduard Trulls, Jan Hosang, Andrea Tagliasacchi, and Kwang~Moo Yi.
\newblock Cotr: Correspondence transformer for matching across images.
\newblock {\em arXiv preprint arXiv:2103.14167}, 2021.

\bibitem{Karel_Vedaldi_BMVC_18}
Karel Lenc and Andrea Vedaldi.
\newblock Large scale evaluation of local image feature detectors on homography
  datasets.
\newblock In {\em The British Machine Vision Conference (BMVC)}, 2018.

\bibitem{li2018megadepth}
Zhengqi Li and Noah Snavely.
\newblock Megadepth: Learning single-view depth prediction from internet
  photos.
\newblock In {\em Proceedings of the IEEE Conference on Computer Vision and
  Pattern Recognition}, pages 3302--3312, 2018.

\bibitem{lindenberger2021pixel}
Philipp Lindenberger, Paul-Edouard Sarlin, Viktor Larsson, and Marc Pollefeys.
\newblock Pixel-perfect structure-from-motion with featuremetric refinement.
\newblock {\em arXiv preprint arXiv:2108.08291}, 2021.

\bibitem{liu2019gift}
Yuan Liu, Zehong Shen, Zhixuan Lin, Sida Peng, Hujun Bao, and Xiaowei Zhou.
\newblock Gift: Learning transformation-invariant dense visual descriptors via
  group cnns.
\newblock {\em arXiv preprint arXiv:1911.05932}, 2019.

\bibitem{DoG}
David~G. Lowe.
\newblock Distinctive image features from scale-invariant keypoints.
\newblock In {\em International booktitle of computer vision}, 2004.

\bibitem{LuoASLFeat2020}
Zixin Luo, Lei Zhou, Xuyang Bai, Hongkai Chen, Jiahui Zhang, Yao Yao, Shiwei
  Li, Tian Fang, , and Long Quan.
\newblock Aslfeat: Learning local features of accurate shape and localization.
\newblock In {\em Proceedings of the IEEE Conference on Computer Vision and
  Pattern Recognition}, 2020.

\bibitem{Melekhov2018}
Iaroslav Melekhov, Aleksei Tiulpin, Torsten Sattler, Marc Pollefeys, Esa Rahtu,
  and Juho Kannala.
\newblock {DGC-Net}: Dense geometric correspondence network.
\newblock In {\em Proceedings of the IEEE Winter Conference on Applications of
  Computer Vision (WACV)}, 2019.

\bibitem{mikolajczykIJCV04}
Krystian Mikolajczyk and Cordelia Schmid.
\newblock Scale \& affine invariant interest point detectors.
\newblock {\em International journal of computer vision}, 60(1):63--86, 2004.

\bibitem{mikolajczyk2005performance}
Krystian Mikolajczyk and Cordelia Schmid.
\newblock A performance evaluation of local descriptors.
\newblock In {\em IEEE Transactions on Pattern Analysis and Machine
  Intelligence}, number~10, pages 1615--1630. IEEE, 2005.

\bibitem{mishchuk2017working}
Anastasiia Mishchuk, Dmytro Mishkin, Filip Radenovic, and Jiri Matas.
\newblock Working hard to know your neighbor's margins: Local descriptor
  learning loss.
\newblock In {\em Advances in Neural Information Processing Systems}, pages
  4826--4837, 2017.

\bibitem{mishkin2015mods}
Dmytro Mishkin, Jiri Matas, and Michal Perdoch.
\newblock Mods: Fast and robust method for two-view matching.
\newblock In {\em Computer Vision and Image Understanding}, volume 141, pages
  81--93. Elsevier, 2015.

\bibitem{mishkin2018repeatability}
Dmytro Mishkin, Filip Radenovic, and Jiri Matas.
\newblock Repeatability is not enough: Learning affine regions via
  discriminability.
\newblock In {\em Proceedings of the European Conference on Computer Vision
  (ECCV)}, pages 284--300, 2018.

\bibitem{goodcorr_kwang}
Kwang Moo~Yi, Eduard Trulls, Yuki Ono, Vincent Lepetit, Mathieu Salzmann, and
  Pascal Fua.
\newblock Learning to find good correspondences.
\newblock In {\em In Proceedings of the IEEE Conference on Computer Vision and
  Pattern Recognition}, pages 2666--2674, 2018.

\bibitem{ono2018lf}
Yuki Ono, Eduard Trulls, Pascal Fua, and Kwang~Moo Yi.
\newblock Lf-net: learning local features from images.
\newblock In {\em Advances in Neural Information Processing Systems}, pages
  6234--6244, 2018.

\bibitem{paszke2017automatic}
Adam Paszke, Sam Gross, Soumith Chintala, Gregory Chanan, Edward Yang, Zachary
  DeVito, Zeming Lin, Alban Desmaison, Luca Antiga, and Adam Lerer.
\newblock Automatic differentiation in pytorch.
\newblock 2017.

\bibitem{Pautrat_2020_ECCV}
Rémi Pautrat, Viktor Larsson, Martin~R. Oswald, and Marc Pollefeys.
\newblock Online invariance selection for local feature descriptors.
\newblock In {\em Proceedings of the European Conference on Computer Vision
  (ECCV)}, 2020.

\bibitem{pautrat2020online}
Rémi Pautrat, Viktor Larsson, Martin~R. Oswald, and Marc Pollefeys.
\newblock Online invariance selection for local feature descriptors.
\newblock In {\em European Conference on Computer Vision}, 2020.

\bibitem{rau2020predicting}
Anita Rau, Guillermo Garcia-Hernando, Danail Stoyanov, Gabriel~J Brostow, and
  Daniyar Turmukhambetov.
\newblock Predicting visual overlap of images through interpretable non-metric
  box embeddings.
\newblock In {\em European Conference on Computer Vision}, pages 629--646.
  Springer, 2020.

\bibitem{revaud2019r2d2}
Jerome Revaud, Philippe Weinzaepfel, C{\'e}sar De~Souza, and Martin
  Humenberger.
\newblock R2d2: Repeatable and reliable detector and descriptor.
\newblock In {\em Advances in Neural Information Processing Systems}, 2019.

\bibitem{Rocco17}
I. Rocco, R. Arandjelovi\'c, and J. Sivic.
\newblock Convolutional neural network architecture for geometric matching.
\newblock In {\em {Proceedings of the IEEE Conference on Computer Vision and
  Pattern Recognition}}, 2017.

\bibitem{sarlin2019coarse}
Paul-Edouard Sarlin, Cesar Cadena, Roland Siegwart, and Marcin Dymczyk.
\newblock From coarse to fine: Robust hierarchical localization at large scale.
\newblock In {\em Proceedings of the IEEE Conference on Computer Vision and
  Pattern Recognition (CVPR)}, 2019.

\bibitem{superglue2019}
Paul-Edouard Sarlin, Daniel DeTone, Tomasz Malisiewicz, and Andrew Rabinovich.
\newblock Superglue: Learning feature matching with graph neural networks.
\newblock In {\em In Proceedings of the IEEE Conference on Computer Vision and
  Pattern Recognition}, 2020.

\bibitem{schonberger2016structure}
Johannes~L Schonberger and Jan-Michael Frahm.
\newblock Structure-from-motion revisited.
\newblock In {\em Proceedings of the IEEE Conference on Computer Vision and
  Pattern Recognition}, pages 4104--4113, 2016.

\bibitem{schonberger2017comparative}
Johannes~L Schonberger, Hans Hardmeier, Torsten Sattler, and Marc Pollefeys.
\newblock Comparative evaluation of hand-crafted and learned local features.
\newblock In {\em Proceedings of the IEEE Conference on Computer Vision and
  Pattern Recognition}, pages 1482--1491, 2017.

\bibitem{shen2020ransac}
Xi Shen, Fran{\c{c}}ois Darmon, Alexei~A Efros, and Mathieu Aubry.
\newblock Ransac-flow: generic two-stage image alignment.
\newblock {\em arXiv preprint arXiv:2004.01526}, 2020.

\bibitem{Simonyan15}
Karen Simonyan and Andrew Zisserman.
\newblock Very deep convolutional networks for large-scale image recognition.
\newblock In {\em International Conference on Learning Representations}, 2015.

\bibitem{sun2021loftr}
Jiaming Sun, Zehong Shen, Yuang Wang, Hujun Bao, and Xiaowei Zhou.
\newblock Loftr: Detector-free local feature matching with transformers.
\newblock In {\em Proceedings of the IEEE/CVF Conference on Computer Vision and
  Pattern Recognition}, pages 8922--8931, 2021.

\bibitem{acne_kwang}
Weiwei Sun, Wei Jiang, Eduard Trulls, Andrea Tagliasacchi, and Kwang Moo~Yi.
\newblock Acne: Attentive context normalization for robust
  permutation-equivariant learning.
\newblock In {\em In Proceedings of the IEEE Conference on Computer Vision and
  Pattern Recognition}, pages 11286--11295, 2020.

\bibitem{d2d2020}
Yurun Tian, Vassileios Balntas, Tony Ng, Axel Barroso-Laguna, Yiannis Demiris,
  and Krystian Mikolajczyk.
\newblock D2d: Keypoint extraction with describe to detect approach.
\newblock In {\em Proceedings of the Asian Conference on Computer Vision
  (ACCV)}, 2020.

\bibitem{tian2020hynet}
Yurun Tian, Axel Barroso-Laguna, Tony Ng, Vassileios Balntas, and Krystian
  Mikolajczyk.
\newblock Hynet: Local descriptor with hybrid similarity measure and triplet
  loss.
\newblock In {\em Advances in neural information processing systems}, 2020.

\bibitem{tian2019sosnet}
Yurun Tian, Xin Yu, Bin Fan, Fuchao Wu, Huub Heijnen, and Vassileios Balntas.
\newblock Sosnet: Second order similarity regularization for local descriptor
  learning.
\newblock In {\em Proceedings of the IEEE Conference on Computer Vision and
  Pattern Recognition}, pages 11016--11025, 2019.

\bibitem{toft2020singleimage}
Carl Toft, Daniyar Turmukhambetov, Torsten Sattler, Fredrik Kahl, and Gabriel
  Brostow.
\newblock Single-image depth prediction makes feature matching easier, 2020.

\bibitem{truong2021gocor}
Prune Truong, Martin Danelljan, Luc~Van Gool, and Radu Timofte.
\newblock Gocor: Bringing globally optimized correspondence volumes into your
  neural network, 2021.

\bibitem{truong2021learning}
Prune Truong, Martin Danelljan, Luc~Van Gool, and Radu Timofte.
\newblock Learning accurate dense correspondences and when to trust them, 2021.

\bibitem{GLUNet_Truong_2020}
Prune Truong, Martin Danelljan, and Radu Timofte.
\newblock {GLU-Net}: Global-local universal network for dense flow and
  correspondences.
\newblock In {\em Proceedings of the IEEE Conference on Computer Vision and
  Pattern Recognition (CVPR)}, 2020.

\bibitem{TuytelaarsMikolajczyk2007}
Tinne Tuytelaars and Krystian Mikolajczyk.
\newblock Local invariant feature detectors: a survey.
\newblock In {\em Foundations and Trends in Computer Graphics and Vision},
  2008.

\bibitem{wiles2020d2d}
Olivia Wiles, Sebastien Ehrhardt, and Andrew Zisserman.
\newblock Co-attention for conditioned image matching.
\newblock In {\em In Proceedings of the IEEE Conference on Computer Vision and
  Pattern Recognition}, 2021.

\bibitem{yi2016lift}
Kwang~Moo Yi, Eduard Trulls, Vincent Lepetit, and Pascal Fua.
\newblock Lift: Learned invariant feature transform.
\newblock In {\em European Conference on Computer Vision}, pages 467--483.
  Springer, 2016.

\bibitem{assignOrient2015}
Kwang~Moo Yi, Yannick Verdie, Pascal Fua, and Vincent Lepetit.
\newblock Learning to assign orientations to feature points.
\newblock {\em CoRR}, abs/1511.04273, 2015.

\bibitem{yu2011asift}
Guoshen Yu and Jean-Michel Morel.
\newblock Asift: An algorithm for fully affine invariant comparison.
\newblock {\em Image Processing On Line}, 1:11--38, 2011.

\bibitem{imwb2020}
Jin Yuhe, Dmytro Mishkin, Anastasiia Mishchuk, Jiri Matas, Pascal Fua, Kwang
  Moo~Yi, and Eduard Trulls.
\newblock Image matching across wide baselines: From paper to practice.
\newblock In {\em International Journal of Computer Vision}, 2020.

\bibitem{zhou2017progressive}
Lei Zhou, Siyu Zhu, Tianwei Shen, Jinglu Wang, Tian Fang, and Long Quan.
\newblock Progressive large scale-invariant image matching in scale space.
\newblock In {\em Proceedings of the IEEE International Conference on Computer
  Vision}, pages 2362--2371, 2017.

\end{thebibliography}
}

\end{document}